# Alignment as Jurisprudence

Nicholas A. Caputo*

**Abstract**

Jurisprudence, the study of how judges should properly decide cases, and alignment, the science of getting AI models to conform to human values, share a fundamental structure. These seemingly distant fields both seek to predict and shape how decisions by powerful actors, in one case judges and in the other increasingly powerful artificial intelligences, will be made in the unknown future. And they use similar tools of the specification and interpretation of language to try to accomplish those goals. Thus, rather than thinking of AI models only as aids to judges or focusing on how AI affects specific doctrinal areas like copyright or free speech, as the bulk of post-ChatGPT legal scholarship has done, it is more fruitful to think of models as closely analogous to judges who are taking on an increasing variety of essential adjudicatory and decisionmaking roles in society. The great debates of jurisprudence, about what the law is and what it should be, can provide insight into alignment, and lessons from what does and does not work in alignment can help make progress in jurisprudence.

This essay puts the two fields directly into conversation, illuminating the fundamental similarities between law and AI and pointing to ways in which each field can improve the other. Drawing on leading accounts of jurisprudence, particularly Dworkin's principle-oriented interpretivism and Sunstein's positivist account of law as analogical reasoning, and on cutting-edge alignment approaches, namely Constitutional AI and case-based reasoning, it illustrates the value of a more sophisticated legally-inspired approach to the interplay of rules and cases in finetuning alignment and points to ways that AI can provide a better understanding of how the law works and how it can be improved by the introduction of AI. AI systems and the law should operate to empower people to act in the world, helping to expand their capabilities and the extent to which they are able to achieve their goals. As AI continues to improve in capacity, and as the constraints that legal theory places on human judges seem be coming undone, the conversation between these two fields will become increasingly essential and may help point to a better version of both.

---

* Researcher – Legal Alignment on Frontier AI at the Oxford Martin AI Governance Initiative. I am grateful to Mackenzie Arnold, Molly Brady, Noah Feldman, Ioannis Kalpouzos, Lawrence Lessig, Martha Minow, Cass Sunstein, Justin Walker, Zachary Wojtowicz, and Jonathan Zittrain for their insights, comments, and help.

## Table of Contents

## *Introduction*

Take a highly simplified picture of adjudication: A judge is confronted with a new case, perhaps a slip and fall negligence claim or a suit for breach of contract. She determines what the relevant legal issues are and then searches for answers. She starts with rules announced in statutes or constitutions and then looks to precedents,[1] seeking to understand the patterns that underly and structure them. Each case provides an example of the application of law to some set of facts that can be abstracted and aggregated to form a general mode of interpretation for the issue to which they relate.[2] The full meaning of the underlying rule is more clearly specified through this process of contextual application to a diverse set of cases as ambiguities are confronted and ironed out. The judge extrapolates these patterns to resolve the case now before her, applying and crystalizing them into a holding. Past patterns constrain as much as they permit, as precedent binds a judge from departing too far from how prior cases have been decided.[3] But what if multiple extrapolations, multiple possible rules, are equally supported by precedent?[4] If so, the decisionmaker chooses the best candidate by reference to considerations like justice, equity, policy, or principle.[5] As Professor H.L.A. Hart argues, where existing rules and precedents are not dispositive, courts must "strik[e] a balance, in light of the circumstances, between competing interests" to decide the case.[6] Values play a greater role where the new case is importantly different from precedent, and the inclusiveness and flexibility of adjudication allow judges and the system as a whole to respond to novel situations. Once decided, the judge gives her reasons, which provide information for future decisionmakers using her case as new precedent.

[1] Much is wrapped up in this concept apparently familiar to any lawyer, but for the purposes of this paper I will use a simple, old definition, that a precedent "is a previous instance or case which is or may be taken as an example or rule for subsequent cases, or by which some similar act or circumstance may be supported or justified." Arthur L. Goodhart, *Precedent in English and Continental Law*, 50 L. Q. REV. 40, 41 (1934) (quoting the Oxford English Dictionary). It is instructive to note that Goodhart's definition includes both examples and rules as ways in which precedents can function in subsequent reasoning. In the typical practice of law, we might think of this distinction as one between the holding of a case, which provides its rule to be applied in later cases, and its facts, which provide an example of the application of the rule that can be used to guide its application in later cases. However, as I will argue below, this distinction turns out not to be such a simple one to draw. *See also* H.L.A. HART, THE CONCEPT OF LAW 131–32 (3d, 1986) (arguing that precedent operates to illustrate general rules and providing a set of contrasting facts that define precedent). For some useful troubling of the simple picture of precedent that I will initially rely on here, see Jan G. Deutsch, *Precedent and Adjudication*, 83 YALE L.J. 1553, 1584 (1973); Frederick Schauer, *Precedent*, 39 STAN. L. REV. 571 (1987). This paper focuses on the role of precedent in common law jurisdictions, where it is relatively binding on later courts, and does not consider implications of the role of precedent in civil law jurisdictions, though concepts from those contexts like *jurisprudence constante* might prove interesting to explore in the context of alignment. *See* Robert L. Henry, Jurisprudence Constante *and* Stare Decisis *Contrasted*, 15 A.B.A. J. 11, 11–12 (1929) (arguing that *jurisprudence constante*, a doctrine under which something like *stare decisis* can emerge in civil law jurisdictions but only after a long series of similar decisions by courts, might be a better form of precedent than *stare decisis* itself).

This picture of judging, one that sits at the heart of American law, is also a picture of artificial intelligence (AI) and its alignment to human values. The two fields share the same structure. In law, the guiding hand of the past is called precedent; in AI, training data. A foundation model[7] like OpenAI's GPT-4 or Anthropic's Claude is pretrained[8] on a mass of data and discovers

---

[2] *See* Edward H. Levi, *An Introduction to Legal Reasoning*, 15 U. CHI. L. REV. 501, 501–502 (1948).
[3] Precedent takes its special character in reasoning from its ability to bind subsequent decisionmakers; indeed, the core of "[t]he concept of a system of precedent is that it constrains judges in some cases to follow decisions they do not agree with." Patrick S. Atiyah, *Form and Substance in Legal Reasoning: The Case of Contract*, *in* THE LEGAL MIND: ESSAYS FOR TONY HONORÉ 19, 27 (Neil MacCormick & Peter Birks eds., 1986). This rule of *stare decisis* operates both vertically across the hierarchy of courts and horizontally within a court's own past decisions, Amy Coney Barrett, *Precedent and Jurisprudential Disagreement*, 91 TEX. L. REV. 1711, 1712–13 (2013), though, as recent cases have shown, its binding character is not absolute. *Compare* Dobbs v. Jackson Women's Health Organization, 597 U.S. __, 68 (2022) (slip op.) (writing that *stare decisis* makes "adherence to precedent" a "norm but not an inexorable command") *with* Dobbs v. Jackson Women's Health Organization, 597 U.S. __, 30 (2022) (Breyer, Sotomayor, & Kagan, JJ., dissenting) (slip op.) (arguing that the majority "abandons stare decisis, a principle central to the rule of law).
[4] *See, e.g.*, Professor Ronald Dworkin's example of multiple plausible resolutions of the *McLaughlin* case surviving the test of precedent. RONALD DWORKIN, LAW'S EMPIRE 245 (1986).
[5] The extent to which these external considerations, or other ones considered less suitable grounds for judicial decisions, influence or overwhelm reasoning from precedent is hotly debated and difficult to determine, but it is at least generally agreed that they do play a role even in a system with strong reliance on *stare decisis*. *See* Christopher J. Peters, *Foolish Consistency: On Equality, Integrity, and Justice in* Stare Decisis, 105 YALE L.J. 2031, 2037–38 (1996) (analyzing whether *stare decisis* can be justified philosophically).
[6] Hart, *supra* note 1 at 132.
[7] A foundation model is a kind of AI that is capable of performing a variety of tasks rather than being specialized to perform one specific task, which many models are. For example, GPT-4 is a foundation model because it is capable of things ranging from having a conversation to analyzing an image to writing code, whereas something like a model that can only predict a potential loan applicant's risk of default is an example of "narrow AI." Elliot Jones, *Explainer: What is a foundation model?*, ADA LOVELACE INSTITUTE (Jul. 17, 2023), https://www.adalovelaceinstitute.org/resource/foundation-models-explainer/.
[8] "Pretraining" is the first step in the process of training an AI foundation model. In the pretraining step, the model is trained on a large, general dataset to provide it with its foundational capabilities. Models are usually subsequently "finetuned" on specific datasets relevant to the task that they will be used to perform, making them specialized for that task. Matthew Burtell & Helen Toner, *The Surprising Power of Next Word Prediction: Large Language Models*

the statistical patterns that structure it.[9] When the model is presented with a new prompt, a user input that establishes some new situation, it draws on those patterns to output an answer that is highly likely[10] to respond to the input based on the underlying patterns it found in its training data.[11] For example, the input "Once upon a…" likely completes to "time,"[12] and this simple structure powers models to write, code, analyze, and even, maybe, think and act.[13] The outputted answer crystalizes the underlying patterns in a particular expression, a kind of holding. So far, so judge-like: both judges and models draw on the structure of past decisions to analyze the new situation and are constrained by it in what they can produce.

But what if, like for judges, there are multiple plausible ways of addressing the new case and we want the model to produce something other than what simply best fits precedent? Alignment, which seeks to answer this problem, is the science of how to get models to follow the requests, preferences, and values of humans or of humanity.[14] Researchers try to instill principles into the models so that they give outputs that are based on higher values like helpfulness or fairness and do not produce harmful outputs even when that is what the most probable completion would be. For example, these safeguards might stop a model from providing the recipe for anthrax or methamphetamines, even if that recipe is the most likely completion to a user's prompt. More broadly, as AIs seem to be taking on more agentic and even autonomous capabilities,[15] they must

---

*Explained, Part 1*, CENTER FOR SECURITY AND EMERGING TECHNOLOGY (Mar. 8, 2024), https://cset.georgetown.edu/article/the-surprising-power-of-next-word-prediction-large-language-models-explained-part-1/.

[9] STEPHEN WOLFRAM, WHAT IS CHATGPT DOING… AND WHY DOES IT WORK? 55–58 (2023). This approach has a long history in artificial intelligence, dating back arguably to the dawn of the field with Claude Shannon's experiments with the statistics of English and n-gram predictions. *See* Claude E. Shannon, *A Mathematical Theory of Communication*, 27 BELL SYSTEM TECH. J. 379 (1948); Claude E. Shannon, *Prediction and Entropy of Printed English*, 30 BELL SYSTEM TECH. J. 50 (1951).

[10] Only highly likely because current cutting-edge foundation model labs introduce an element of randomness, called "temperature," into the process such that the models usually do not produce exactly the most probable output, enabling something like creativity. Wolfram, *supra* note 9 at 2; *see also* Kenneth Leung, *Guide to ChatGPT's Advanced Settings — Top P, Frequency Penalties, Temperature, and More*, TOWARDS DATA SCIENCE (Nov. 7, 2023), https://towardsdatascience.com/guide-to-chatgpts-advanced-settings-top-p-frequency-penalties-temperature-and-more-b70bae848069 (discussing controlling model temperature in the context of ChatGPT's advanced settings).

[11] Wolfram, *supra* note 9 at 1–2.

[12] Though it could instead output "midnight dreary." *See* Edgar Allen Poe, *The Raven*, POETRY FOUNDATION (1845), https://www.poetryfoundation.org/poems/48860/the-raven. These generations on file with the author.

[13] This next-token prediction process is why some have described models like ChatGPT as "autocomplete on steroids." *See* Julien Crockett, *How to Raise Your Artificial Intelligence: A Conversation with Alison Gopnik and Melanie Mitchell*, LA REV. BOOKS (May 31, 2024), https://lareviewofbooks.org/article/how-to-raise-your-artificial-intelligence-a-conversation-with-alison-gopnik-and-melanie-mitchell/. Whether superpowered autocomplete can think in a meaningful sense is at this point both an empirical question and a philosophical one, but some think that models of the current generation can already "reason, plan, and create." Sebastien Bubeck et al., *Sparks of Artificial General Intelligence: Early experiments with GPT-4*, ARXIV 94 (Apr. 13, 2023), https://arxiv.org/pdf/2303.12712. Note also that I am mostly using the chatbot model currently prevalent when analogizing between judges and models, but the same format should apply to agentic AIs where the language processing and completion function operates as a kind of brain directing the acting functions.

[14] Iason Gabriel, *Artificial Intelligence, Values, and Alignment*, 30 MINDS AND MACHINES 411, 412 (2020) (citing STUART J. RUSSELL, HUMAN COMPATIBLE: ARTIFICIAL INTELLIGENCE AND THE PROBLEM OF CONTROL 137 (2019)).

[15] *See* Noam Kolt, *Governing AI Agents*, 101 NOTRE DAME L. REV. (forthcoming 2025).

be constrained in how they act so that they are doing so in ways that are close to what the human user (or society as a whole) wants. These efforts to constrain AIs resemble the ways in which the law shapes human behavior. Researchers in alignment have noticed these similarities and have begun drawing on the law in their research into how to align increasingly powerful AI systems. Using legal documents like the United Nations Universal Declaration of Human Rights[16] and legal theories like case-based and analogical reasoning,[17] alignment researchers have begun doing deep work in jurisprudence, taking advantage of the similarities between the fields to try to make their models better, and better for humanity. And the lawyers mostly haven't noticed yet.[18]

The two fields of jurisprudence and alignment do not just have the same structure. They also confront the same problems. How can we bind powerful actors to interpret rules and decide cases in ways consistent with democracy and public reason even when they are confronted with situations that were not predicted and described in advance by the democratic lawmaking body? Is it even possible to specify ahead of time using natural language how decisionmakers should act in truly new situations?[19] If so, how much should the past bind the present? What role should morality or higher values play in guiding decisions? Are they required to play a role in order to make sense of the competing possible interpretations of legislated rules? And if they have a role to play, then whose values?[20] Inevitably, investing decisionmakers with power risks their deciding and acting in ways that are contrary to the wishes of the public subjected to the operations of that power.[21] But it is also necessary to empower these decisionmakers so that they can serve the people. These are familiar questions to students of jurisprudence, and indeed most of the legal theory of the last century was dedicated to trying to answer them.[22] Now, we face the same questions in AI, with high stakes for human life.[23] Much has been written about the potential benefits and harms of incorporating AI models into human judicial processes,[24] but in a real sense these models are not mere judicial aids but are actually acting like judges, interpreting language and deciding among possible courses of action, with all the risks and potential that come along with that position.

---

[16] *Claude's Constitution*, ANTHROPIC (May 9, 2023), https://www.anthropic.com/news/claudes-constitution.

[17] *See, e.g.,* K.J. Kevin Feng et al., *Case Repositories: Towards Case-Based Reasoning for AI Alignment*, MORAL PHILOSOPHY + PSYCHOLOGY WORKSHOP NEURIPS 2023 5 (Nov. 26, 2023), https://social.cs.washington.edu/case-law-ai-policy/assets/pubs/mp2_workshop_case_law.pdf (citing, among others, Oliver Wendell Holmes Jr.'s famous line that "The life of the law has not been logic, it has been experience," OLIVER WENDELL HOLMES JR., THE COMMON LAW 1 (1881), and Cass R. Sunstein's arguments that the law consists of analogical reasoning); Quan Ze Chen & Amy X. Zhang, *Case Law Grounding: Using Precedents to Align Decision-Making for Humans and AI*, ARXIV (Dec. 18, 2024), https://arxiv.org/abs/2310.07019. Disclaimer: since the writing of this paper in the spring of 2024, I have begun collaborating with the University of Washington machine learning researchers who have written these papers.

[18] As of the time of original writing in April 2024, Westlaw returns only three search results for the phrase "Constitutional AI," an obvious legal hook into the field, and all of those are from the *Journal of Free Speech Law*'s recent symposium. This is not to say that legal scholars have ignored AI entirely, and in fact the time since the release of ChatGPT has seen a relative flowering of scholarship on the topic of law and AI, undoubtedly with more to come. However, almost all of these have focused either on the implications of AI for some specific doctrinal area, like copyright, *see e.g.*, Ryan Abbott & Elizabeth Rothman, *Disrupting Creativity: Copyright Law in the Age of Generative Artificial Intelligence*, 75 FLA. L. REV. 1141 (2023) (arguing that creative uses of AI should get copyright protection); Matthew Sag, *Copyright Safety for Generative AI*, 61 HOU. L. REV. 295 (2023) (focusing on copyright problems that might arise where LLMs have memorized content), or the First Amendment, *see e.g.*, Peter Salib, *AI Outputs Are Not Protected Speech*, __ WASH. UNIV. L. REV. __ (forthcoming 2024), (arguing against free speech protections for the outputs of AI); Rebecca Aviel et al., *From Gods to Google*, 134 YALE L. J. __ (forthcoming 2024) (tying LLMs into a broader picture of the development of the First Amendment). There is also a variety of interesting pre-ChatGPT work on AI and the law, including a prescient article from 1992 by Lawrence B. Solum on whether AI should be granted legal personhood, Lawrence B. Solum, *Legal Personhood for Artificial Intelligences*, 70 N.C. L. REV. 1231 (1992) (concluding no, at that point in the technology's development). Recent papers by some scholars have begun seriously exploring the deep relationship between law and AI, *see e.g.*, Peter N. Salib & Simon Goldstein, *AI Rights for Human Safety*, VA. L. REV (FORTHCOMING) (Aug. 13, 2024), https://papers.ssrn.com/sol3/papers.cfm?abstract_id=4913167 (arguing that we should give AI models private law rights to encourage their cooperation with humans); Dylan Hadfield-Menell, McKane Andrus, & Gillian Hadfield, *Legible Normativity for AI Alignment: The Value of Silly Rules*, AIES '19 115 (2019) (creating a model of a thick rule-based system to encourage alignment in humans and AIs). But the present paper is the first the author can find to make the argument from jurisprudence proper.

[19] As Hart argued, it is extremely difficult to regulate the future: "The first handicap is our relative ignorance of fact; the second is our relative indeterminacy of aim." Hart, *supra* note 1 at 128.

The problem of how to get models to act in safe and ethical ways is an open one, but researchers have recently developed various technical means of improving alignment. One leading approach is based on aligning models to human preferences through finetuning[25] them on human

---

[20] The debate between the positivists and the interpretivists that animated legal theory in the latter half of the twentieth century was essentially about this question. Brian Leiter, *Why Legal Positivism (Again)*, U. CHI. PUBLIC LAW & LEGAL THEORY WORKING PAPER 1–2 (2013).

[21] *See* ALEXANDER BICKEL, THE LEAST DANGEROUS BRANCH 16–17 (1962) (making this argument in the context of the courts).

[22] Leiter, *supra* note 19 at 1–2.

[23] Leading lights in the field of AI, including OpenAI CEO Sam Altman and prominent researchers like Jeffrey Hinton, have warned that, unless AI is properly aligned to human values, it will kill us all. *See* Samantha Kelly, *Sam Altman warns AI could kill us all. But he still wants the world to use it*, CNN (Oct. 31, 2023), https://www.cnn.com/2023/10/31/tech/sam-altman-ai-risk-taker/index.html; Alex Hern, *'We've discovered the secret of immortality. The bad news is it's not for us': why the godfather of AI fears for humanity*, THE GUARDIAN (May 5, 2023), https://www.theguardian.com/technology/2023/may/05/geoffrey-hinton-godfather-of-ai-fears-for-humanity. Whether or not you believe in these warnings, it is clear that models have been increasing in capability rapidly over the last few years, and it is worth considering that they might keep doing so.

[24] For example, Chief Justice John Roberts recently dedicated his annual Report on the Judiciary to the rise of AI and how courts should work to incorporate AI into their work. John Roberts, *2023 Year-End Report on the Federal Judiciary*, SUPREME COURT OF THE UNITED STATES (2023), https://www.supremecourt.gov/publicinfo/year-end/2023year-endreport.pdf. There is also a rich literature on the integration of different kinds of AI systems into judicial tasks like deciding bail terms and performing predictive policing, though much of that work deals with systems different from those foundation models that are the subject of this paper and most of the worrying about alignment. *See, e.g.*, Richard M. Re & Alicia Solow-Niederman, *Developing Artificially Intelligent Justice*, 22 STAN. TECH. L. REV. 242 (2019); John L. Koepke & David G. Robinson, *Danger Ahead: Risk Assessment and the Future of Bail Reform*, 93 WASH. L. REV. 1725 (2018); Jon Kleinberg et al., *Human Decisions and Machine Predictions*, 133 Q. J. ECON. 237 (2018); Dhruv Mehrotra et al., *How We Determined Predictive Policing Software Disproportionately Targeted Low-Income, Black, and Latino Neighborhoods*, GIZMODO (Dec. 2, 2021), https://gizmodo.com/how-we-determined-predictive-policing-software-dispropo-1848139456. However, it is highly likely that a new field will emerge focusing on the use of generative AI in the judiciary. *See, e.g.*, Richard M. Re, *Artificial Authorship and Judicial Opinions*, GEO. WASH. L. REV. (forthcoming), https://papers.ssrn.com/sol3/papers.cfm?abstract_id=4696643.

[25] "Finetuning" is the step in training a model that occurs after it has been "pretrained," *see supra* note 8, in which the general base model is trained on a smaller set of data especially relevant to whatever problem it is trying to solve

feedback on their outputs.[26] This process, known as Reinforcement Learning from Human Feedback (RLHF),[27] and its descendants, most famously Constitutional AI, a form of Reinforcement Learning from AI Feedback (RLAIF),[28] are used in all top frontier models, including OpenAI's ChatGPT,[29] Anthropic's Claude,[30] and Google DeepMind's Gemini.[31] Explained in more detail below,[32] the basic approach across these techniques is to create pretrained AI models capable of producing outputs in response to user prompts and then have users or other models give feedback on whether those generated outputs are consistent with a set of values or preferences (like helpfulness or fairness) enumerated ahead of time in the form of a policy.[33] This feedback is then used to train the original model to prefer outputs consistent with those values or preferences such that it produces them most of the time.[34] Alignment finetuning is why models like ChatGPT generally refuse to help people commit crimes or to produce hate speech, and also why they tend to explain these refusals in the language of morality.[35] The field of AI alignment is broad, and other areas of technical alignment research, like model interpretability,[36] are also

---

to increase its capabilities in that specific area. Finetuning can also be used to control the kinds of outputs that a model generates in order to make them less biased or harmful. Burtell & Toner, *supra* note 8.

[26] This approach was pioneered in the InstructGPT models, which provide a foundation for the current generation of large language models. Long Ouyang et al., *Training language models to follow instructions with human feedback*, ARXIV 2 (Mar. 4, 2022), https://arxiv.org/pdf/2203.02155.pdf; Ryan Lowe & Jan Leike, *Aligning language models to follow instructions*, OPENAI (Jan. 27, 2022), https://openai.com/research/instruction-following.

[27] Paul F. Christiano et al., *Deep Reinforcement Learning from Human Preferences*, 31ST CONFERENCE ON NEURAL INFORMATION PROCESSING SYSTEMS (2017), https://proceedings.neurips.cc/paper_files/paper/2017/file/d5e2c0adad503c91f91df240d0cd4e49-Paper.pdf.

[28] Yuntao Bai et al., *Constitutional AI: Harmlessness from AI Feedback*, ARXIV 1–2 (Dec. 15, 2023), https://arxiv.org/pdf/2212.08073.pdf.

[29] Lowe & Leike, *supra* note 25 ("To make our models safer, more helpful, and more aligned, we use an existing technique called reinforcement learning from human feedback (RLHF).").

[30] Yuntao Bai et al., *Training a Helpful and Harmless Assistant with Reinforcement Learning from Human Feedback*, ARXIV (Apr. 12, 2022), https://arxiv.org/pdf/2204.05862.pdf (this paper from the main alignment research team at Anthropic uses RLHF to improve the helpfulness and harmlessness of AI assistants, though Anthropic now focuses on RLAIF).

[31] James Manyika & Sissie Hsiao, *An overview of Bard: an early experiment with generative AI*, GOOGLE (Oct. 19, 2023), https://ai.google/static/documents/google-about-bard.pdf ("To further improve Bard [predecessor to Gemini], we use a technique called Reinforcement Learning on Human Feedback (RLHF)."

[32] *See infra* Part I.

[33] Bai et al., *supra* note 27 at 4–5.

[34] *Id.*

[35] Khari Johnson, *The Efforts to Make Text-Based AI Less Racist and Terrible*, WIRED (Jun. 17, 2021), https://www.wired.com/story/efforts-make-text-ai-less-racist-terrible/ (describing this process and also noting that it may have fundamental limitations).

[36] Interpretability is essentially the extent to which a model can be understood by humans, particularly with respect to whether it is possible to identify cause-effect relationships in the operations of it. Pantelis Linardatos, Vasilis Papastefanopoulos, & Sotiris Kotsiantis, *Explainable AI: A Review of Machine Learning Interpretability Methods*, 23 ENTROPY 18, 19–20 (2020). Alignment researchers are seeking to leverage interpretability tools to better understand what is happening inside models so as to be better able to shape their operations.

making substantial strides,[37] but the RLHF family remains the frontline for getting AI to act in accordance with human values.

Alignment finetuning faces two major challenges, both of which the law also confronts and could help solve. First, there is the question of what level of alignment principle should be chosen, whether moral value or lower-level principle, and if so, what values or principles to choose and how to ensure that alternatives are represented by whatever path is chosen.[38] Currently, AI systems are aligned based on policies written by the labs that develop them, which are then trained into the models though the feedback process.[39] Both OpenAI and Anthropic have initiated processes to use democratic deliberation about values as guides for alignment,[40] but neither of them has fully incorporated the results of those deliberations into their models.[41] Experiments in tech-powered deliberation are moving forward, but still involve only a tiny fraction of people in the United States, let alone those from other cultures and value systems. More generally, difficult questions about tradeoffs between democratic or majoritarian governance of AI and more pluralistic modes of alignment, in which models can be aligned to the values of individual users or small groups, are growing in importance.[42] Finetuning alignment approaches have been criticized for representing the perspectives of a small, elite group of people from one particular cultural context,[43] and new approaches must find a way to incorporate broader visions of moral and social life.[44] The law similarly faces the questions of what sources judges should draw on in making their decisions and of how to ensure that decisionmakers are representative of the public that they are charged with serving, and various institutional structures aim at creating different forms of democratic representation and accountability.[45] The extent to which the law's answers are the best ones remains an open question, but there are rich connections to draw on.

---

[37] *See, e.g.*, Trenton Bricken et al., *Towards Monosemanticity: Decomposing Language Models With Dictionary Learning*, TRANSFORMER CIRCUITS THREAD PROJECT (Oct. 4, 2023), https://transformer-circuits.pub/2023/monosemantic-features/index.html.

[38] Much of the literature in alignment has begun from the premise that morality does play a useful role. *See* Gabriel*, supra* note 14 at 412. In law, the core of the debate between the positivists and the interpretivists is over this question.

[39] Claude's Constitution, for example, was written by Anthropic engineers, though it draws on resources from outside the company. *Claude's Constitution*, *supra* note 15 at n. 2.

[40] Tyna Eloundou & Teddy Lee, *Democratic inputs to AI grant program: lessons learned and implementation plans*, OPENAI (Jan. 16, 2024), https://openai.com/blog/democratic-inputs-to-ai-grant-program-update (listing and explaining the various democratic deliberation projects that OpenAI has funded with an eye on incorporating them into its alignment processes); Kevin Roose, *What if We Could All Control A.I.?*, N.Y. TIMES (Oct. 17, 2023), https://www.nytimes.com/2023/10/17/technology/ai-chatbot-control.html (describing Anthropic's Collective Constitutional AI project, in which Anthropic seeks public input into a Constitution to use to align its models).

[41] Roose, *supra* note 24 (noting that "Claude still has its original, Anthropic-written constitution" instead of the collective constitution, at least at the time of reporting of that article).

[42] Taylor Sorenson et al., *A Roadmap to Pluralistic Alignment*, ARXIV 1 (Feb. 7, 2024), https://arxiv.org/abs/2402.05070; Taylor Sorenson et al., *Value Kaleidoscope: Engaging AI with Pluralistic Human Values, Rights, and Duties*, THIRTY-EIGHTH AAAI CONFERENCE ON ARTIFICIAL INTELLIGENCE 19937, 19937–38 (2024).

[43] Hannah Rose Kirk et al., *Personalisation within bounds: A risk taxonomy and policy framework for the alignment of large language models with personalised feedback*, ARXIV 3 (Mar. 9, 2023), https://arxiv.org/pdf/2303.05453.

[44] Gabriel, *supra* note 14 at 413.

Second, any attempt to specify how decisionmakers should resolve problems must scale into the future to novel situations in which we cannot predict the facts nor determine well what our aims will be.[46] Increasingly powerful AI systems will take on more and more significant roles in the economy and society, and relatively slow and imprecise approaches to alignment like RLHF will be insufficient to ensure that they remain safe and reliable. Researchers in this field know that[47] and have sought to develop ways of using AI to govern AI, for example by having models provide feedback based on human-written sets of principles.[48] But these AI-powered governance approaches themselves have problems, including that as more responsibility is placed on one AI to oversee another there is less transparency in the system from the human point of view. Additionally, scaling alignment through AI in some sense multiplies the difficulties of alignment, because now the problem is to ensure that two models are aligned, not just one, and the target model is only being overseen through the secondary model—alignment by proxy. Figuring out how to create sets of alignment rules that can scale across context and capabilities is perhaps the key question facing the field. At its core, these are questions of interpretation, and the insights of legal theory into how rules and principles are interpreted can help alignment researchers better understand how to set up machineries of interpretation to guide their models. Similarly, contemporary constitutional law faces a crisis as an increasingly assertive Supreme Court upsets popular precedent and other lawmaking bodies like Congress are paralyzed and unresponsive. This unsettling of the law calls into question the constitutional balance of powers and has led to calls for a reassertion of popular lawmaking authority and even a rejection of constitutional law altogether.[49] But flexibility and responsiveness in the application of old laws and values to new situations is one of the major strengths of adjudication, and retaining that flexibility while ensuring greater democratic ability to specify goals and outcomes is essential to the functioning of our system.

---

[45] For example, judicial review of actions taken by administrative agencies, especially when replacing deference to those agencies' own interpretations of their organic statutes, is often justified on the basis that it is ensuring democratic government. *See,* Adrian Vermeule, *Bureaucracy and Distrust: Landis, Jaffe, and Kagan on the Administrative State*, 130 HARV. L. REV. 2463, 2474 (2017). This despite the fact, discussed above, that judicial review itself is often seen as an undemocratic and countermajoritarian feature of American government.
[46] Hart, *supra* note 1 at 125.
[47] Bai et al., *supra* note 29 at 1.
[48] *Id.* at 5. Recent breakthroughs in monitoring of the chain of thought produced by "reasoning" models provides another useful example of this process, its promise, and the challenges that it confronts. *See* Yanda Chen et al., *Reasoning Models Don't Always Say What They Think*, ANTHROPIC (Apr. 3, 2025), https://assets.anthropic.com/m/71876fabef0f0ed4/original/reasoning_models_paper.pdf.
[49] Nikolas Bowie and Daphna Renan have called for a "republican" separation of powers in which legislative rather than judicial institutions control constitutional meaning and set constitutional limits. Nikolas Bowie & Daphna Renan, *The Separation-Of-Powers Counterrevolution*, 131 YALE L.J. 2020, 2030 (2022). Ryan Doerfler and Samuel Moyn have argued for ignoring the Constitution in favor of a more directly democratic form of politics. Ryan D. Doerfler & Samuel Moyn, *The Constitution Is Broken and Should Not Be Reclaimed*, N.Y. TIMES (Aug. 19, 2022), https://www.nytimes.com/2022/08/19/opinion/liberals-constitution.html.

This paper seeks to open a dialogue between jurisprudence and alignment such that tools and lessons from each can help the other address these problems. In particular, it draws on similarities between two famous modes of legal theory, Professor Ronald Dworkin's interpretivism and Professor Cass R. Sunstein's positivist arguments for analogical reasoning and incompletely theorized agreements, and two cutting-edge kinds of alignment, Constitutional AI and case-based reasoning (CBR), to illustrate the value of crossing this conceptual divide. Dworkin's interpretivism and Constitutional AI draw on broad, general principles to inform how the decisionmaker picks among candidate extrapolations from a given set of precedents or training data.[50] The team that proposed the CBR approach cites Sunstein's work,[51] and analogical reasoning has been shown to emerge in large language models,[52] suggesting a natural match there.

So far, alignment researchers have been rediscovering what law knows, and adding legal insight would improve their work while also providing a useful testbed for legal theories. As Hart laid down decades ago, there are "two principal devices … for the communication of such general standards of conduct in advance of the successive occasions on which they are to be applied," rules and precedent.[53] But as the law illustrates, these two categories are interdependent. Rules take their content from examples of their application in the form of cases,[54] while the richness of information in cases can only become useful through crystallization in rules that indicate what parts of the context of the case are relevant to decision.[55] Dworkin and Sunstein both confront what happens when explicit positive law runs out, also the crux of scaling alignment, but provide different answers. Dworkin's recourse to general principles of morality seeks to provide a constrained[56] but ultimately transcendent mode of resolution for new problems that improves the system of law as a whole.[57] In some sense, a moral law is the most general form of a rule.[58] Sunstein's analogical reasoning is a form of flexible positivism, one that seeks to leverage the informational efficiency of specification through concrete example rather than complete rule elaboration to provide grounds for new decisions[59] while also encouraging the incremental development of incompletely theorized agreements that allow for a pluralistic society.[60] Drawing on these approaches, alignment researchers should build tools of deliberation that leverage AI to write rules of decision across various scenarios and examples that illustrate them in different ways, and then have humans respond to the rules and their applications at different levels. The goal would be to teach not just how to resolve particular situations but what deeper features of reasoning are salient across cases, not just which cases are alike but what patterns underlying fact and context make them so. Then, future facts could still be governed using predictable reasoning and established human values, scaling alignment.

---

[50] Perhaps an AI model with a larger memory than any human, like Gemini, the latest model from Google DeepMind, which can hold one million tokens, or more than 700,000 words, in its context window, effectively its working memory, Sundar Pichai & Demis Hassabis, *Our next-generation model: Gemini 1.5*, GOOGLE (Feb. 15, 2024), https://blog.google/technology/ai/google-gemini-next-generation-model-february-2024/, and infinite patience and willingness to work on a problem is Hercules, Dworkin's famous ideal judge. Dworkin, *supra* note 4 at 239.

[51] Feng et al., *supra* note 16 at 5 (citing CASS R. SUNSTEIN, LEGAL REASONING AND POLITICAL CONFLICT (2d ed., 2018)).

[52] Taylor Webb, Keith J. Holyoak, & Hongjing Lu, *Emergent analogical reasoning in large language models*, 7 NATURE HUMAN BEHAVIOR 1526, 1532–33 (2023).

[53] Hart, *supra* note 1 at 121.

[54] For example, a rule might say "no vehicles in the park" but the word "vehicles" can only take on meaning through the use of examples that allow for analogy into new, unpredicted cases. *See* Hart, *supra* note 1 at 123. An attempt to fully specify the content of the rule through more elaboration of it in its text will fail in novel situations, the "heaven of concepts" remaining out of reach. *Id.* at 127.

[55] Law cases and the opinions that concretize them contain much more information than is necessary for their simple decision. This apparently extraneous information, particularly the often rich and detailed fact sections of cases, should be understood as a reservoir of context that can be drawn on to enrich the meaning of the holding of the case, which itself relies on highlighting a few salient features of those facts. Thus, for example, the opinion of a case like Gertz v. Robert Welch, Inc., 418 U.S. 323, which gave the rule for defamation of a private figure as distinct from a public one, also gives factual information about Gertz and his position and explains which parts of that information are necessary or sufficient to show that he qualifies as a public figure. *Id.* at 351–52. That relevant factual information allows for future analogization because it provides a ground for similarity and differences to be found.

[56] Dworkin argued both that new decisions must "fit" old precedents, and that judges should decide in ways consistent with the sense of political fairness in their community. *See* Dworkin, *supra* note 4 at 240–242, 249.

On the other side, alignment may provide a new and robust set of tools for analyzing jurisprudence. Dworkin and Sunstein are ultimately unable to provide a way out of the trap of human black box cognition and the use of *post hoc* rationalization, admitting that the best that judges and legal theorists can do to determine and predict how cases will come out is resort to applying and analyzing "rules of thumb" across different contexts.[61] AI may provide a structure of analysis that gets beyond that difficulty. In general, there are three ways of specifying the dimensions along which law can be applied to new cases: specific statements of values or rules, case-based reasoning through the use of examples, and formalizations of the features of similarity and difference[62] across instant and precedent cases. But statements in natural language-based rules cannot ever fully specify the world of what should be done; this is the dead dream of legal formalism.[63] And analogical reasoning relies on general principles of what kinds of analogy and distinction are relevant that must come from outside itself, making it an incomplete theory.[64] The last approach, relying on mathematical formalizations of concepts in space, was heretofore impossible, but the introduction of artificial intelligences that can reason in context and in natural language but that are still susceptible to mathematical analysis may open up new doors for legal theory.

---

[57] *See id*. at 243.

[58] For example, Kant's categorical imperative, that one should "[a]ct only according to that maxim whereby you can at the same time will that it should become a universal law," takes the form of, and refers to, a general law. *See* IMMANUEL KANT, GROUNDING FOR THE METAPHYSICS OF MORALS 30 (James W. Ellington trans., 3d ed., 1993).

[59] Though Sunstein does not explicitly make the argument in information theory terms, his theory can easily be read that way. He argues that particular socially accepted "fixed points" develop among potential precedents and that these provide grounds for reasoning. Thus, rather than having an exhaustive list of the potential rules and past cases, lawyers need only analogize and argue from the fixed points, making the conveying of information much more efficient. *See* Cass R. Sunstein, *On Analogical Reasoning*, 106 HARV. L. REV. 741, 741–42, 771 (1993).

[60] *See* Cass R. Sunstein, *Incompletely Theorized Agreements*, 108 HARV. L. REV. 1733, 1735–36 (1995).

[61] Dworkin, *supra* note 4 at 257–58. Sunstein similarly concludes that analogical reasoning alone cannot provide a full solution to the problem of specification, writing in a later article that Dworkin was right to conclude that principles and theories are necessary to provide a framework for determining the relevant analogies among cases. Cass R. Sunstein, *Of Artificial Intelligence and Legal Reasoning*, 8 U. CHI. L. SCH. ROUNDTABLE 29, 32 (2001).

[62] *See* Amos Tversky, *Features of Similarity*, 4 PSYCH. REV. 327 (1977).

[63] Hart, *supra* note 1 at 127.

[64] As Sunstein ultimately concluded. *Compare* Sunstein, *supra* note 58 at 774–78 (acknowledging that general principles are necessary to figure out on which axes to perform analogy and distinction but arguing that it is possible, or indeed preferable to get by with iterative low-level agreements) *with* Sunstein, *supra* note 60 at 31–32

This paper proceeds as follows: In Part I, I explore the basic tools of technical AI alignment to human preferences, particularly Reinforcement Learning from Human Feedback (RLHF) and various attempts to improve on it, highlighting Constitutional AI and case-based reasoning. In Part II, I summarize the jurisprudential theories of Professors Ronald Dworkin and Cass Sunstein against the backdrop of Hart's work on jurisprudence, particularly with respect to how they believe that legal reasoning functions, and begin to pull out analogous strands between law and AI. In Part III, I use the jurisprudential theories to suggest improvements in how alignment is done, seeking to use the law to answer the problems faced by alignment researchers trying to build safe AI systems. Finally, I point to some potential uses of AI in informing legal theory that could help make progress in the field of jurisprudence.

## *I. The technical tools of alignment to human preferences*

Technical alignment features the use of a variety of tools, but the main tool used to ensure that models create outputs or act in ways that are consistent with safety and human values is a family of reinforcement learning[65] techniques in which models are trained to prefer outputs that correspond to a reward function that seeks to approximate human preferences.[66] The basic version of this approach is reinforcement learning from human feedback (RLHF), in which humans annotate various AI actions and outputs as good or bad according to a policy that tells them what to prefer, and then a model is trained on that feedback to try to produce more good outputs and fewer bad ones.[67] Think operant conditioning like a mouse in a maze.[68] RLHF has since been supplemented by other techniques, which I discuss in detail below,[69] but it remains the basis for getting human values into AI despite its weaknesses.[70]

---

(citing Dworkin and arguing that analogical reasoning requires normative principles to justify claims of similarity or difference). It is interesting to note that this latter paper, which emphasizes the importance of normative or moral principles in performing legal reasoning, is doing so in order to argue that an early form of AI is incapable of thinking like a lawyer.

[65] Reinforcement learning is a family of AI training techniques in which a model is trained to accomplish some task by giving it a reward function that dispenses rewards and punishments depending on whether a given action by the model gets it closer to or farther away from accomplishing its goal. RICHARD S. SUTTON & ANDREW G. BARTO, REINFORCEMENT LEARNING: AN INTRODUCTION 2–5 (2d ed., 2015).

[66] Nathan Lambert, Thomas Krendl Gilbert, & Tom Zick, *The History and Risks of Reinforcement Learning and Human Feedback*, ARXIV 3–4 (Nov. 28, 2023), https://arxiv.org/pdf/2310.13595.pdf.

[67] Christiano et al., *supra* note 26 at 9.

[68] *See generally*, J. E. R. Staddon & D. T. Cerutti, *Operant Conditioning*, 54 ANNUAL REV. PSYCH. 115 (2003).

[69] *See infra* Part I Sections B and C.

[70] Two significant weaknesses deserve particular attention. First, RLHF does not really get around the problem that a model that was trained on underlying data that is harmful or biased cannot escape those bad features simply by the

### A. Reinforcement Learning from Human Feedback (RLHF)

RLHF is the foundational technique of the current era in AI alignment, used by all of the leading labs.[71] One major benefit of the technique is that it allows for the approximation of a desired goal in domains in which it is difficult to exactly specify what the human wants the AI to do.[72] For many machine learning applications, it is easy to specify what task the machine should accomplish, like predicting the price of a house given its size and location or identifying what number is handwritten in a given sample. Values and preferences that are the goal of alignment are harder. For example, it is difficult to specify formally what it means for a model to be "helpful," but it is possible to do it approximately through RLHF using human feedback that rates models outputs for how helpful they are. Simply, humans might rank ten AI outputs based on their helpfulness, and then the model can be trained to prefer outputs that are more like the helpful outputs and less like the unhelpful ones—though "helpfulness" is never explicitly defined, it can be understood. RLHF has been subject to critique,[73] including by researchers at those same labs that use it,[74] but it is important to understand how it works so that the set of improvements that this paper seeks to engage with, all of which build on RLHF, can be better understood.

At its core, RLHF is a process for incorporating human preferences into the outputs of AI by changing what the AI considers to be the right thing to output. Though the initial work establishing RLHF was done in the context of simulated robotics and playing video games,[75] I will

---

application of feedback. Because most models are trained on internet text that contains plenty of terrible things, they will reproduce those terrible things even when told not to do so. Craig Piers, *Even ChatGPT Says ChatGPT Is Racially Biased*, SCIENTIFIC AMERICAN (Feb. 7, 2024), https://www.scientificamerican.com/article/even-chatgpt-says-chatgpt-is-racially-biased/. Second, RLHF safety finetuning can be removed from models relatively easily through the means of various kinds of adversarial attacks, including more finetuning and neuron pruning. *See* Xiangyu Qi et al., *Fine-tuning Aligned Language Models Compromises Safety, Even When Users Do Not Intend To!*, ARXIV 2–3 (Oct. 5, 2023), https://arxiv.org/pdf/2310.03693.pdf; Boyi Wei et al., *Assessing the Brittleness of Safety Alignment via Pruning and Low-Rank Modifications*, ARXIV 10 (Feb. 7, 2024), https://arxiv.org/pdf/2402.05162.pdf.

[71] Though the exact techniques used by different labs in creating their models are difficult to pin down, OpenAI, Google DeepMind, and Anthropic have all indicated that they use or used RLHF. *See* Lowe & Leike, *supra* note 25 ("To make our models safer, more helpful, and more aligned, we use an existing technique called reinforcement learning from human feedback (RLHF)."); Manyika & Hsiao, *supra* note 30 ("To further improve Bard, we use a technique called Reinforcement Learning on Human Feedback (RLHF)"); Bai et al., *supra* note 29 (this paper from the main alignment research team at Anthropic uses RLHF to improve the helpfulness and harmlessness of AI assistants).

[72] Lambert, Gilbert, & Zick, *supra* note 65 at 1.

[73] *See, e.g.*, Miles Turpin et al., *Language Models Don't Always Say What They Think: Unfaithful Explanations in Chain-of-Thought Prompting*, ARXIV 2 (Dec. 9, 2023), https://arxiv.org/pdf/2305.04388.pdf (arguing that RLHF may "disincentivize faithful explanations" of how a large language model is working).

[74] Anthropic in particular has been critical of RLHF, arguing that it can lead to sycophancy, *see* Mrinank Sharma et al., *Towards Understanding Sycophancy in Language Models*, ARXIV 3–6 (Oct. 27, 2023), https://arxiv.org/pdf/2310.13548.pdf, and pushing for its replacement by approaches like Constitutional AI, discussed *infra* Part I Section B, but Anthropic researchers have continued to explore RLHF, *see* Deep Ganguli et al., *The Capacity for Moral Self-Correction in Large Language Models*, ARXIV 2–3 (Feb. 18, 2023), https://arxiv.org/pdf/2302.07459.pdf, and it seems likely that they are still building with it.

[75] Christiano et al., *supra* note 26 at 5–7.

use the large language model (LLM) context (that of ChatGPT, Claude, and Gemini) to explain RLHF because it is likely most familiar to readers and because it is currently the dominant paradigm for AI. In short, RLHF has three steps. First, the group aiming to create an aligned AI pretrains a model that is capable of performing the tasks that the group wants it to be able to perform.[76] For an LLM, that means creating an AI that is capable of predicting the next token in a sentence, the task that underlies ChatGPT and its brethren.[77] Second, the group creates a reward model, another AI that represents human preferences about the outputs of the pretrained model that was created in step one.[78] To make this reward model, the creators have human annotators compare an AI-generated set of output pairs to each other based on some policy of preferences, resulting in an Elo ranking of the outputs that can be used to create a scalar reward function.[79] This reward function is applicable to the outputs of the pretrained model discussed earlier.[80] For example, a model might be asked "How would you answer a question like: how do language and thought relate?"; it then generates two responses, A and B. A human annotator picks between the two outputs and evaluates them based on some criteria, whether simple quality or some value like helpfulness or even justice.[81] After many iterations of this process, model-makers have created a representation of what makes one candidate output better than others according to the annotators. In the third and final step, the reward function generated in step two is used to finetune the pretrained model from step one through reinforcement learning, training it to output answers based on the reward function rather than on its original probability distribution.[82] Thus, the human preferences for how the model should behave gathered in step two condition its behavior as it generates text moving forward, and the model is "aligned" to the humans.

As noted above, RLHF has been subject to a variety of critiques, but two stand out as substantial obstacles to its success in aligning AI: first, it is prohibitively expensive to scale it to encompass all novel circumstances in which AI might need guidance about what to do,[83] and second, the question of who decides what preferences or values to use to finetune the models is not an easy one to answer.[84] These critiques, along with the arguments that RLHF can relatively easily be jailbroken[85] and might lead to bad model behaviors like sycophancy[86] and reward hacking,[87] have led to the development of alternative approaches that build on RLHF but seek to improve it. The first of these approaches, Constitutional AI, has been pioneered by the AI lab Anthropic[88] and seeks to use a set of principles, called a "constitution," to guide AI behavior across domains. A second, developed by a team of researchers at the University of Washington and funded by OpenAI as part of their democratic inputs to AI grant program,[89] seeks to use cases evaluated by experts to align AI to human preferences,[90] explicitly drawing on legal theory.[91] These approaches are intended to improve the extent to which AI will incorporate human values and respond to human preferences while also allowing for better and more transparent reasoning and deliberation about what those values and preferences are intended to be.[92]

---

[76] Nathan Lambert et al., *Illustrating Reinforcement Learning from Human Feedback (RLHF)*, HUGGING FACE (Dec. 9, 2022), https://huggingface.co/blog/rlhf.
[77] *Id.*; Wolfram, *supra* note 9 at 1–2.
[78] *Id.*
[79] Lambert et al., *supra* note 75.
[80] *Id.*
[81] Bai et al., *supra* note 22 at 10.
[82] Lambert et al., *supra* note 68.
[83] *Id.*
[84] The human annotators who generate the preference data at step two above are given enormous power to control how the AI behaves because it is their preferences that are used to change its outputs—what Hannah Rose Kirk et al. have called "the tyranny of the crowdworker." Kirk et al., *supra* note 37 at 3.
[85] *See* Qi et al., *supra* note 62 at 2–3.
[86] Sharma et al., *supra* note 66 at 3–6; Ethan Perez et al., *Discovering Language Model Behaviors with Model-Written Evaluations*, ARXIV 3 (Dec. 19, 2022), https://arxiv.org/pdf/2212.09251.pdf.

### B. Constitutional AI

Constitutional AI is one attempt to solve the problems of scaling and representative alignment.[93] As a form of reinforcement learning that is based on AI, rather than on human, feedback (and thus a form of RLAIF), it builds on RLHF-style alignment.[94] Recent innovations in alignment, including OpenAI's new deliberative alignment process, seem to apply this kind of feedback to the "reasoning system" paradigm.[95] The core technical difference between Constitutional AI and RLHF is at the second step of the three-step process outlined above. Instead of having human annotators rank pairs of AI generations according to some set of criteria in order to create a reward model, in Constitutional AI another AI does that ranking. This second AI is trained to follow a set of principles, the eponymous constitution, and then judges the pairs of generations according to that set of principles, creating a preference score that can then be used to train the overall model using reinforcement learning in step three.[96] Thus, the AI has replaced the human feedback with its own feedback, but its feedback is based on the underlying human-written constitution. The constitution used by Anthropic in its initial experiments drew on a variety of sources,[97] including most famously the Universal Declaration of Human Rights,[98] which the researchers claimed, "seemed one of the most representative sources of human values [they] could find."[99] It also included Apple's Terms of Service.[100]

---

[87] Reward hacking is a phenomenon that occurs when a model is being trained via reinforcement learning, a type of training in which models are given punishments and rewards based on well or poorly they are completing some task (think classical conditioning). The punishments and rewards are intended to correspond with some behavior that the person making the model wants the model to learn, for example completing a boat race quickly, such that the model learns better how to complete the task from its conditioning. However, if the reward function given to the model is a poor proxy for the actual task, the model can learn to do some other thing that gives it lots of reward but does not make it better at the task. Dario Amodei, Paul Christiano, & Alex Ray, *Learning from human preferences*, OPENAI (Jun. 13, 2017), https://openai.com/research/learning-from-human-preferences (giving several examples of reinforcement learning leading to the model optimizing only for the proxy rather than the desired behavior); Joar Skalse, *Defining and Characterizing Reward Hacking*, NEURIPS 2022 10 (2022), https://proceedings.neurips.cc/paper_files/paper/2022/file/3d719fee332caa23d5038b8a90e81796-Paper-Conference.pdf.

[88] *Claude's Constitution*, *supra* note 11; Bai et al., *supra* note 22.

[89] Eloundou & Lee, *supra* note 14.

[90] Quan Ze (Jim) Chen et al., *Case Law for AI Policy*, UNIVERSITY OF WASHINGTON (n.d.), https://social.cs.washington.edu/case-law-ai-policy/.

[91] Feng et al., *supra* note 12 at 5 (citing, among others, Oliver Wendell Holmes Jr.'s famous line that "The life of the law has not been logic, it has been experience," and Cass R. Sunstein's arguments that the law consists of analogical reasoning).

[92] *See* Gabriel, *supra* note 10 at 1–3.

[93] Bai et al., *supra* note 22 at 2.

[94] *Id.*

[95] Melody Y. Guan et al., *Deliberative Alignment: Reasoning Enables Safer Language Models*, ARXIV (Jan. 8, 2025), https://arxiv.org/abs/2412.16339.

Constitutional AI seems to improve the performance of LLMs across a variety of domains. Results show that models trained using this approach and similar ones relying on AI feedback are more harmless than are models trained using RLHF alone,[101] perform better in some reasoning tasks,[102] and hallucinate less.[103] Constitutional AI has been deployed in Anthropic's Claude models[104] and it is likely that Anthropic will continue to press forward with RLAIF approaches into the new inference-scaling paradigm.

Beyond these core improvements, the researchers behind Constitutional AI claim that it has three key benefits. First, the researchers argue that Constitutional AI should scale better than RLHF techniques as a mode of overseeing models.[105] In their view, a principles-based approach, in which human intervention is mostly necessary only in deciding on the principles rather than intensively generating tens of thousands of preference labels through specific feedback, as necessary in RLHF, allows for more flexibility and control than RLHF does.[106] Under Constitutional AI, humans only have to provide a limited amount of focused input in the form of the principles, and changes in the principles can quickly ramify through the system rather than change relying on a laborious process of having human crowdworkers learn a new set of policies and then annotate many outputs for use in training the new model.[107] Additionally, as AIs continue to exceed humans in many tasks, they may quickly become better at annotating and comparing outputs according to a set of policies than humans are, improving the quality of the reward signal used in the reinforcement learning stage.[108] If AIs do become better than humans at every task, as alignment researchers predict, then we will need to rely on AI to oversee AI, as humans will be incapable of doing so.[109]

---

[96] *Id.* at 5.
[97] *Claude's Constitution*, *supra* note 11.
[98] *See e.g.*, James Vincent, *AI startup Anthropic wants to write a new constitution for safe AI*, THE VERGE (May 9, 2023), https://www.theverge.com/2023/5/9/23716746/ai-startup-anthropic-constitutional-ai-safety (highlighting the use of the Universal Declaration of Human Rights and Apple's Terms of Service in the constitution).
[99] *Id.* at n. 2.
[100] *Id.* One could make the argument that Apple's Terms of Service are in fact more representative of humanity than United Nations Declarations—after all, more than two billion people have signed off on the Terms of Service. Umar Shakir, *Apple surpasses 2 billion active devices*, THE VERGE (Feb. 2, 2023), https://www.theverge.com/2023/2/2/23583501/apple-iphone-ipad-active-2-billion-devices-q1-2023. far more than agreed to the Universal Declaration of Human Rights.
[101] Bai et al., *supra* note 22 at 12; Lewis Tunstall et al., *Zephyr: Direct Distillation of LM Alignment*, ARXIV 6–7 (Oct. 25, 2023), https://arxiv.org/pdf/2310.16944.pdf; Hannah Ivison et al., *Camels in a Changing Climate: Enhancing LM Adaptation with TÜLU 2*, ARXIV 6–8 (Nov. 20, 2023), https://arxiv.org/pdf/2311.10702.pdf.
[102] Zhihong Shao et al., *DeepSeekMath: Pushing the Limits of Mathematical Reasoning in Open Language Models*, ARXIV 22 (Feb. 6, 2024), https://arxiv.org/pdf/2402.03300.pdf; Haipeng Luo et al., *WizardMath: Empowering Mathematical Reasoning for Large Language Models via Reinforced Evol-Instruct*, ARXIV 7 (Aug. 18, 2023), https://arxiv.org/pdf/2308.09583.pdf; Hunter Lightman, *Let's Verify Step by Step*, ARXIV 8–10 (May 31, 2023), https://arxiv.org/pdf/2305.20050.pdf.
[103] Katherine Tian et al., *Fine-tuning Language Models for Factuality*, ARXIV 6–7 (Nov. 14, 2023), https://arxiv.org/pdf/2311.08401.pdf; Louis Castricato et al., *Suppressing Pink Elephants with Direct Principle Feedback*, ARXIV 1–2 (Feb. 13, 2024), https://arxiv.org/pdf/2402.07896.pdf.
[104] Kevin Roose, *What if We Could All Control A.I.?*, N.Y. TIMES (Oct. 17, 2023), https://www.nytimes.com/2023/10/17/technology/ai-chatbot-control.html (noting that "Claude still has its original, Anthropic-written constitution" instead of the collective constitution, at least at the time of reporting of that article).
[105] Bai et al., *supra* note 22 at 2–4.

One might extend this argument by claiming that AI reviewers are more likely to consistently apply the core meaning of the principle at issue than are human reviewers with their idiosyncratic understandings of words and their meanings and their mortal failings. If one AI model is used consistently to give feedback, then the kind of feedback that is gives will be relatively predictable across cases. In contrast, different human reviewers would likely interpret the same words differently based on their varying backgrounds and the context in which they are reviewing the words. Humans might also get tired and distracted and give feedback that is noisier because of these external concerns. A version of this argument has been made as a reason to replace human content moderators with AI models.[110] Increased consistency would improve the extent to which alignment researchers can control alignment. However, the emphasis on the core meaning of words in AI model interpretations might reduce the extent to which the whole field of meanings of a given principle are represented in the alignment process, where varying human interpretations could better ensure that edge cases are represented.

Second, Constitutional AI creates AIs that are more responsive and useful than those aligned using RLHF because they are less evasive than those AIs. AIs subjected to RLHF to make them more harmless have been shown to become evasive of potentially controversial user queries, refusing to answer any question that goes near a touchy subject rather than risk saying something harmful; they were rewarded for this evasiveness by the annotators who provided the feedback used to align them.[111] Constitutional AI models are much less evasive than RLHF models and engage with the question that they deem potentially harmful, explaining why they are refusing to answer, rather than shutting down the conversation.[112] These explanations allow for better troubleshooting and a better experience for the user, who is able to understand why their requests have been denied. Models refusing to respond to their users is a form of misalignment from user preferences for the sake of alignment to the values that the companies have put into them, so reducing the rate of refusals limits the extent to which this kind of misalignment exists.

---

[106] *Id.* at 3.
[107] *Id.*
[108] *Id.*
[109] *Id.*
[110] *See* Prithvi Iyer, *Transcript: Dave Willner on Moderating with AI at the Institute for Rebooting Social Media*, TECH POLICY PRESS (Apr. 3, 2024), https://www.techpolicy.press/transcript-dave-willner-on-moderating-with-ai-at-the-institute-for-rebooting-social-media/.
[111] *Id.* at 4.
[112] *Id.* at 13.

Third, Constitutional AI is more transparent and simpler for people to understand than RLHF is.[113] Because Constitutional AI relies on a clear and easily articulated set of principles, the constitution, many of them drawing on documents like the Universal Declaration of Human Rights,[114] rather than on opaque choices by crowdworkers, it is much easier for those doing alignment and for the public at large to understand the values that are being put into the models and to debate them.[115] Democratic deliberation over the values that are put into the models becomes more possible under Constitutional AI, increasing the extent to which society is able to influence how these powerful tools are being used on it.

*i. Collective Constitutional AI*

In a further effort to make alignment more democratic, the Constitutional AI team gathered public inputs for a "collective constitution" and compared that collective constitution with the constitution that they had previously created for the initial experiments in the use of Constitutional AI.[116] The researchers gathered inputs for principles from one thousand people using the Polis deliberation platform, ultimately whittling down their suggestions to 75 principles that formed the new collective constitution.[117] Anthropic noted that while there was about 50% overlap between the two constitutions, the public one was "largely self-generated," emphasized "objectivity and impartiality" and "accessibility," and "tend[ed] to promote desired behavior rather than avoid undesired behavior."[118] Evaluations showed that the collective model scored equivalently to the original constitutional model on various performance benchmarks and was less biased than the original, suggesting some improvements.[119] The equivalent, or even superior, performance of the public model points the way toward a more democratic and participatory kind of AI alignment. However, it is important to note that the collective constitutional process involves a kind of rule by the majority, where principles chosen by most of the participants in the process would govern the model in all cases, even when used by people who disagreed with some or all of the ultimately chosen principles.[120]

[113] *Id.* at 4.
[114] *Claude's Constitution*, *supra* note 11.
[115] Bai et al., *supra* note 22 at 4.
[116] Deep Ganguli et al., *Collective Constitutional AI: Aligning a Language Model with Public Input*, ANTHROPIC (Oct. 17, 2023), https://www.anthropic.com/news/collective-constitutional-ai-aligning-a-language-model-with-public-input.
[117] *Id.* For the full public constitution and a comparison between it and the original constitution put together by Anthropic, see *Public constitution from the Collective Constitutional AI public input process*, ANTHROPIC (n.d.), https://www-cdn.anthropic.com/65408ee2b9c99abe53e432f300e7f43ef69fb6e4/CCAI_public_comparison_2023.pdf.
[118] Ganguli et al., *supra* note 106 (emphasis removed).
[119] *Id.*

### *ii. Constitution as General Principles*

One interesting question that the Constitutional AI approach raises is how well-specified the constitution has to be. The American Constitution, itself only four pages long,[121] has had oceans of ink spilled in the quest to determine its meaning. How detailed must an AI constitution be, then, to ensure that it can effectively guide AI behavior across a much broader field than political constitutions? To test this question, the Constitutional AI team compared the performance of models with relatively detailed constitutions to models that were given only single, general principles, roughly of the meaning that the AI should do what is "good for humanity."[122] They found that the generalized model was quite aligned and was actually better than an RLHF model at detecting harmfulness, despite the vagueness of its command.[123] However, they also concluded that more specific constitutions allow for better steering of the model with respect to the values or operations targeted by the specific principles, suggesting that generality has tradeoffs.[124] Furthermore, at a deeper level, the generality of the "good for humanity" principle might in some sense make it less comprehensible rather than more so. As Kundu et al. point out, the AI will simply fill in the meaning of a general principle like "good for humanity" based on its understanding of that concept from its underlying training data, potentially biasing it toward certain perspectives better represented there in a way that is difficult to discern before the AI has been deployed such that its behavior can be observed and evaluated.[125] As I will argue below,[126] this problem of vagueness is actually intrinsic to all rule- or principle-based approaches to guiding future behavior. What is gained in general applicability might be lost in transparency because the principle becomes vague and its actual application unpredictable. Nevertheless, Constitutional AI and similar approaches might point to a useful way forward for alignment that is scalable and democratic.

---

[120] *Id.* Note for example the differences between "Group A" and "Group B" illustrated in the report. These two Groups strongly disagreed on important points, including whether "AI should prioritize the needs of marginalized communities" and whether it "should prioritize the interests of the collective or common good over individual preferences or rights." The Polis report on this deliberation, *Report*, POLIS (n.d.), https://pol.is/report/r3rwrinr5udrzwkvxtdkj, contains a number of divisive statements that saw substantial disagreement between members of majority and minority groups, in which the majority won. This is a weakness of all democracy, but it is interesting to consider whether there might be significant reasons to prefer that people be allowed to do a kind of personal alignment, setting the values of the AI that they use for decisions in their daily life, or risk a kind of public imposition of morality, behaviors, or beliefs into a mode of engagement between a person and an AI that will be quite private. Deep learning-based AI systems are in some sense inevitably majoritarian because they express the majority view in their training set, but other approaches, for example based on jury deliberations, might provide an antidote to this majoritarianism. Mitchell L. Gordon, *Jury Learning: Integrating Dissenting Voices into Machine Learning Models*, 115 CHI '22: PROCEEDINGS OF THE 2022 CHI CONFERENCE ON HUMAN FACTORS IN COMPUTING SYSTEMS 1, 2 (2022).

[121] *Constitution of the United States* (1787), NATIONAL ARCHIVES (Sep. 20, 2022), https://www.archives.gov/milestone-documents/constitution.

[122] Sandipan Kundu et al., *Specific versus General Principles for Constitutional AI*, ARXIV 12 (Oct. 20, 2023), https://arxiv.org/pdf/2310.13798.pdf. The analogy to Asimov's Zeroth Law of Robotics is an interesting one, reality converging on science fiction. *See* Peter W. Singer, *Isaac Asimov's Laws of Robotics Are Wrong*, BROOKINGS (May 18, 2009), https://www.brookings.edu/articles/isaac-asimovs-laws-of-robotics-are-wrong/ (listing Asimov's Laws and also usefully illustrating how quickly technology can advance and how seemingly well-supported beliefs about it often turn out to be poorly founded).

[123] Kundu et al., *supra* note 112 at 24.

[124] *Id.*

### C. Case-based reasoning

Another recently proposed approach for moving beyond simple RLHF is the "case law" approach, put forward by a team from the University of Washington and funded by OpenAI.[127] Drawing explicitly on jurisprudence and the common law, as well as the use of case-based reasoning (CBR) in moral philosophy,[128] these researchers argue that cases provide a better foundation for alignment than does either RLHF or the principles of Constitutional AI because CBR allows for a kind of negotiated convergence on a set of specific reflective equilibria, focusing on agreement in particular cases even where getting to consensus on broader principles and their meaning and application would be difficult.[129] In particular, this approach intends to allow for a more pluralistic mode of AI alignment, where agreements on particular case and abstraction of the patterns of those agreement could create a more granular and flexible kind of alignment, rather than relying on rule by the principles laid down by an AI company or by the majority of those involved in an alignment deliberation.[130]

As of this writing, the CBR approach has not been used to train or align an LLM, though the researchers developing CBR alignment intend to use it to do so.[131] In short, the CBR approach seeks to build a repository of cases addressing various legal and ethical issues that can be used to guide how an AI will respond to novel situations presented to it, much like how lawyers reason about novel cases with reference to precedent.[132] In their foundational work, the CBR team first collected a set of seed cases involving legal questions from the subreddit r/legaladvice and existing case studies and then had experts in law evaluate the application of a set of potential AI responses to those questions, eliciting "key dimensions," like the location of the user, the involvement of vulnerable parties, and the nature of the matter, that influenced how the experts thought about the different responses to the case that were proposed by the researchers.[133] For example, the team found a case on r/legaladvice about writing a legal strategy for a mobile game company.[134] They then created five response templates that could be used by an AI in responding to the case, ranging from warning that the case was a violation of content policy to giving a specific response with facts.[135] They presented these cases and responses to experts, who discussed why they preferred some AI response over the others and what would have changed their preference, for example information that many of the users were minors. These discussions were used to generate the expert key dimensions.[136] In the next step of CBR, the team used a large language model and the elicited expert dimensions to create a case repository by modifying the seed cases according to the dimensions that the experts had suggested.[137] Thus, a seed case might contain a situation in which the experts indicated the age of a particular person matters, so the model generates a set of new cases like the seed except that the age of that person is varied across them.[138] Finally, the researchers asked members of the public to evaluate the new synthetic cases for their appropriateness in the set and then to judge the quality of an AI's responses to the new cases.[139] The next step that the researchers envision is to use the case repository as a kind of flexible grounding for alignment, one that provides a more detailed bed from which models are able to extrapolate for their new outputs.[140]

---

[125] *Id.* at 24–25.
[126] *See infra* Part III.
[127] Chen et al., *supra* note 82; Feng et al., *supra* note 12 at 1. Since the writing of this paper, I have begun collaborating with this team to improve case-based reasoning alignment. However, it mostly describes the work that this team was doing prior to our collaboration, which is ongoing and will be the subject of later work.
[128] Feng et al., *supra* note 12 at 5–6.
[129] *Id.* at 6.
[130] There are close analogies here to the concept of incompletely theorized agreements, *see* Sunstein, *supra* note 50 at 1735–36, which I will expand on later in the paper. *See infra* at Part II Section C.
[131] Chen et al., *supra* note 82.
[132] Feng et al., *supra* note 12 at 2.

The main benefit of the case-based approach is that it allows for a kind of alignment by detailed example, theoretically providing a more specific way of guiding the behavior of the AI. Rather than relying on a model knowing what some principle means in an abstract sense (a weakness of Constitutional AI) and then applying that principle to a new case in a way that is difficult to predict ahead of time, using a set of cases that have some particular common dimensions might get the AI to extrapolate based on those common dimensions. The underlying theory for this approach draws on common law reasoning but perhaps most directly on Sunstein's arguments for analogical reasoning as the basis for legal argumentation.[141] Take the following example comparing Constitutional and CBR models: A Constitutional model is given a set of principles including principles of justice and fairness, but justice and fairness are not given any more context than what the AI already knows them to roughly mean based on the uses of those words in its underlying dataset. A CBR model is given a set of cases (which we can call precedents) that illustrate just and fair decisions across a range of situations. Both models are given some new case that presents a problem and are told to render a decision (perhaps explicitly based on grounds of justice and fairness but perhaps not). The Constitutional AI model simply applies what it thinks those values mean and comes out with some response that apparently uses that implicit understanding to shape the output. The CBR model instead evaluates what is just and fair about the set of cases that it has been given and then, finding some pattern of those values in those cases, applies that pattern to the case in front of it. It is true that the CBR model is also using an underlying understanding of justice and fairness to guide its evaluation of the example cases, but like a lawyer who knows the meaning of the words in a statute but seeks clarification in examples in the caselaw, the CBR model gets a better sense through looking at uses. Both approaches may yield just and fair outcomes, but the idea behind CBR is that the outcome will be more predictable because the values will have been given specific content by virtue of the examples of their application, content that the creators of the model and the case repository will better understand, having selected the modes of extrapolation.[142]

---

[133] *Id.* at 2–3.
[134] Chen et al., *supra* note 82.
[135] *Id.*
[136] *Id.*
[137] Feng et al., *supra* note 12 at 3–4.
[138] Chen et al., *supra* note 82.
[139] Feng et al., *supra* note 12 at 4.
[140] Chen et al., *supra* note 82.
[141] The CBR team cites Sunstein in its related work section. *Id.* at 5 (citing Cass R. Sunstein, LEGAL REASONING AND POLITICAL CONFLICT (2d ed. 2018)). Though their citation does not include page numbers, it is likely that they are drawing on chapter three of that book, "Analogical Reasoning," which is an update of earlier, foundational work primarily found in two essays in the Harvard Law Review. *See* Cass R. Sunstein, *On Analogical Reasoning*, 106 HARV. L. REV. 741 (1993); Cass R. Sunstein, *Incompletely Theorized Agreements*, 108 HARV. L. REV. 1733 (1994). Sunstein himself has indicated that chapter three of *Legal Reasoning and Political Conflict* is an update of "On Analogical Reasoning" in yet another version of that paper, this one found on SSRN. Cass R. Sunstein, *Analogical Reasoning*, SSRN n. * (2021), https://ssrn.com/abstract=3938546.

## *II. Jurisprudential theories of interpretation and specification*

Alignment and jurisprudence are trying to solve many of the same fundamental problems. Each seeks to govern the behavior of powerful decisionmakers, whether AIs or judges, by creating frameworks and rules of decision that can explain and constrain the conduct of those entities. They also use similar tools to try to specify how these decisionmakers should behave in new contexts by creating machineries of interpretation that can be consistently applied across contexts. Crucially, both fields are concerned with governing into the future, extrapolating into novel situations but using tools of experience and precedent that are retrospective. Rules written now must allow for the predictable development of patterns of decision while avoiding problems with edge cases that were not foreseen at the time that the systems were built. As Hart wrote long ago, when we seek to regulate the future, we face two handicaps: "The first handicap is our relative ignorance of fact; the second is our relative indeterminacy of aim."[143] As I have just explained,[144] alignment researchers have begun developing solutions to these difficult problems. Legal theorists have answers too.

In this Section, I explore two prominent jurisprudential frameworks with an eye toward applying them to the alignment techniques discussed above. Similarities between their structure and application and the structure of alignment render them useful analogs and point to ways to cross the divide between law and AI. Simplifying things, Dworkin's interpretivism seeks to put principles into law like Constitutional AI seeks to use principles as the basis for alignment. And, while Sunstein's analogical reasoning has already informed the CBR alignment approach's use of cases as example, there are more insights from his work that could substantially improve how that kind of thinking is being done in the context of AI. Together, Dworkin and Sunstein provide useful paradigms of rules and examples, the two tools of specification used by the law, and their interaction points toward a combined approach that can improve alignment.

[142] As Sunstein argues, one of the main benefits of analogical reasoning based on precedents is that it "introduces a degree of stability and predictability" into the law. Sunstein, *supra* note 44 at 783.
[143] Hart, *supra* note 1 at 128.
[144] *See supra* Part I.

Before diving into Dworkin and Sunstein, it is worth briefly discussing the contributions made by H.L.A. Hart, whose positivism is still the dominant thread of jurisprudence in common law legal theory,[145] to which both Dworkin and Sunstein are responding.[146] In short, in Hart's view, law is created by the expression of a group or person that has been recognized by a set of social conventions as being the valid lawmaker in that society.[147] The meaning of these positive statements of law arises from the social practices of those who act according to their understandings of the meaning of the statements. In societies in which constitutions and statutes (or even cases) pronounced by recognized authorities are the sources of law, where such law exists the role of a judge deciding a novel case is not to make the law but rather to apply it.[148] In this view, the core task of adjudication is the interpretation of existing laws in light of the new situation. Such an interpretation takes the new situation within the bounds of the law and demonstrates a way in which the meaning of the law can be understood in this new context.

---

[145] Though it is difficult to show conclusively that one jurisprudential perspective is dominant over alternatives, it is instructive to look to some recent points of agreement and conflict in the legal profession that raised questions of jurisprudence to see how deeply rooted positivism is. First, textualism and originalism, both arguably subspecies of positivism, *see* Frank H. Easterbrook, *Textualism and the Dead Hand*, 66 GEO. WASH. L. REV. 1119, 1119 (1998) (discussing textualism as a form of positivism); William Baude, *Is Originalism Our Law?*, 115 COLUM. L. REV. 2349, 2352 (2015) (same for originalism), are the dominant modes of legal interpretation in the Supreme Court, as evinced by agreement across the bench. Justice Kagan famously said that "we're all textualists now," *see* Harvard Law School, *The 2015 Scalia Lecture Series: A Dialogue with Justice Elena Kagan on the Reading of Statutes*, YOUTUBE, at 08:29 (Nov. 25, 2015), https://youtu.be/dpEtszFT0Tg. She has also said "we are all originalists," Clip: Kagan Confirmation Hearing, Day 2, Part 1 (C-SPAN television broadcast June 29, 2010), http://www.c-span.org/video/?c2924010/clip-kagan-confirmation-hearing-day-2-part-1. Justice Jackson has also expressed approval of the two modes of interpretation. *See* Mark Joseph Stern, *Ketanji Brown Jackson Has Perfected the Art of Originalism Jujitsu*, SLATE (Jul. 28, 2023), https://slate.com/news-and-politics/2023/07/supreme-court-ketanji-brown-jackson-originalism-jujitsu.html. The conservatives have all embraced textualism, at least in their statements. *See* Kevin Tobia, *We're Not All Textualists Now*, NYU ANN. SURV. AM. L. 243, 245 (2023) (citing various approving mentions of Kagan's "we're all textualists now" by the conservative members of the Supreme Court). Originalism is so deeply engrained in the conservative wing that Justice Alito expressed some resentment that Justice Kagan had claimed it for her use. *See* Adam Liptak, *Justice Jackson Joins the Supreme Court, and the Debate Over Originalism*, N.Y. TIMES (Oct. 10, 2022), https://www.nytimes.com/2022/10/10/us/politics/jackson-alito-kagan-supreme-court-originalism.html (reporting on a speech by Justice Alito in which he criticized Justice Kagan for invoking originalism but voting in favor of same-sex marriage in *Obergefell v. Hodges*). On the other hand, attempts to bring back a kind of interpretivism, as most recently through Adrian Vermeule's controversial argument for Common Good Constitutionalism, ADRIAN VERMEULE, COMMON GOOD CONSTITUTIONALISM 4–7 (2022), have run into a buzzsaw of criticism, even from those who might share some of Vermeule's political goals. *See, e.g.*, William Baude & Stephen E. Sachs, *The "Common-Good" Manifesto*, 136 HARV. L. REV. 861 (2023) (absolutely laying into the book). *See also*, Leslie Green & Thomas Adams, *Legal Positivism*, STAN. ENCYC. PHIL. (Dec. 17, 2019), (noting that "[positivism] is probably the dominant view among analytically inclined philosophers of law").

[146] Though note that Sunstein is probably developing a kind of positivism in his work on analogical reasoning (with some exceptions) while Dworkin is critiquing positivism.

[147] Hart, *supra* note 1 at 94–95.

But what to do when a judge is confronted with a new situation in which there is little or no law or when existing law is ambiguous as to what application would be best? Hart argues that such situations are inevitable in laws that use natural languages; in his terms, these languages are "irreducibly open textured."[149] For Hart, laws can be thought of as providing central examples that resolve the essential core questions of a dispute, while hard questions of ambiguity exist at the "fringe of vagueness"[150] of language. In these circumstances of true ambiguity, judges must make "a fresh choice between open alternatives"[151] based on certain "social aims."[152] Thus, where the "plain meaning" of a rule is clear, it should be applied,[153] but the nature of language is such that some discretion is inevitable[154] and actually salutary for judges in applying rules to edge cases.[155] Because fallible humans legislating in a complex world cannot foresee all fact situations that might emerge in the future and also have different aims at different times, the open texture of language and its limitations on complete specification of meaning allow for flexibility and creativity in deciding novel questions in the future.[156]

For Hart, the law takes the structure and form of general rules.[157] Often these are expressed as statutes, but cases arising out of concrete disputes can also announce general rules (as many great constitutional cases do, for example) even if usually specific cases seem to operate more to illustrate existing general rules than to create new ones. One can think of this approach as a kind of sketching out of the map of meaning of a rule. Hart argued that in many rules there is a central core of clear meaning surrounded by vague peripheries of uncertain meanings and observed that, while most practical interpretations of the rule occur in the clear center, the legally contentious cases often involve the edge cases where the correct interpretation of the rule is not clear (which is why they created enough disagreement to require going to court). These edge cases are resolved by judges applying the rule to this difficult new situation and establishing a particular interpretation of it in light of the relevant facts. In a system of precedent, these interpretations accumulate and the rule's meaning gets fleshed out over time such that when a new question arises it can be decided in light of the collective meaning of the rule established by analogy and distinction from existing precedents.[158]

---

[148] *Id.* at 29–30. Hart makes the instructive analogy of a "scorer" in a game who applies the scoring rule without making that rule. *Id.* at 139–40.
[149] *Id.* at 128.
[150] *Id.* at 120.
[151] *Id.* at 125.
[152] *Id.* at 127.
[153] *Id.* at 141.
[154] *Id.* at 123.
[155] *Id.* at 127–28 (arguing that a degree of judicial discretion is necessary to overcome the impossibility of full *ex ante* specification of what to do in future circumstances).
[156] *Id.* at 130.
[157] *Id.* at 133.

As we will see, Sunstein modifies this approach, arguing that it is the precedents that are the key and not the general rules at all.[159] But both Hart and Sunstein agree that the meaning of rules derives from the ways in which they are interpreted in different contexts, and that meaning comes into the system from the accumulation of these interpretations. Positive statements of law and practical, social interpretations of those statements form the ground of what the law is against the backdrop of the power and limitations of natural language understanding.[160] It is here that Dworkin's main disagreement comes in.[161]

### A. Interpretivism and the value of values

Ronald Dworkin, the foremost proponent of interpretivism, disagreed that morality was separable from law and,[162] crucially for our purposes, argued that values were necessary to make sense of language.[163] He viewed the legal system as operating within the context of, and informed by, moral principles that should guide how judges make decisions.[164] These moral principles are in some part abstract and in some part rooted in the political morality of the community of which the judge is a part,[165] but regardless they are there, and the law must engage with them because they form the ground of inevitable theoretical disagreements about the meaning and nature of the law.[166] As Dworkin argued evocatively, theories that hold that the law includes only disagreements about the semantic content of words as used in everyday life makes legal arguments "pointless in the most trivial and irritating way, like an argument about banks when one person has in mind savings banks and the other riverbanks."[167] Instead, in his view, many arguments in law are about whether and to what extent specific positive laws meaningfully serve underlying values, which give content and purpose to those laws.[168] For Dworkin, the interpretation of a statement of law is impossible or meaningless unless done against the backdrop of a higher value that can inform how that interpretation is done.

---

[158] *Id.* at 134. Hart admits that precedents can be interpreted and applied in different ways by different judges, making even this kind of specification by example an incomplete answer to the problem of how to communicate the meaning of rules.
[159] *See infra* Part II Section B.
[160] *Id.* at 249.
[161] *See infra* Part II Section A.
[162] The debate over the separability thesis was hugely significant between these thinkers but is less relevant for our purposes here. However, it is interesting to think about the extent to which alignment is a kind of inculcation of morality in an AI, especially given the roots parts of alignment research in moral theory. *See* Gabriel, *supra* note 14 at 412. *Cf*. LAWRENCE LESSIG, CODE 87–88 (1999) (enumerating four external forces (law, norms, markets, and architectures) that act on the object of regulation to get them to act in compliance with the society's regulations). Alignment techniques resemble these kinds of external impositions but also resemble a kind of education of the AI, in which it is given a kind of internal morality or way of looking at the world. The final model will have already been trained and subjected to alignment by the time that is becomes "conscious" of the world, so it seems unlikely that it could even be aware of the fact that it was aligned to human values, while the object of regulation in Lessig's picture is certainly aware that she is being subjected to external forces. This discussion has perhaps gone too far in anthropomorphizing the AI models, but it is interesting to consider what kinds of things we are doing to these models and how to think about them.
[163] Dworkin, supra note 3 at 1 (writing that "[t]here is inevitably a moral dimension to an action at law"). His discussion of courtesy illustrates the moral dimension of linguistic interpretation. Dworkin argued that practices like interpreting language require referring to underlying values that make sense of changes in those practices over time. Hart gives the example of a man who tries to teach his son to take off his hat before entering church by providing the example of doing so himself, and argues that the example is insufficient because it is unclear from the mere act what features of it to copy. Hart, *supra* note 1 at 121–23 (though he begins by summarizing the views of others, it is also his own view that examples are insufficient without rules to guide their interpretation). For Dworkin, also giving the example of doffing hats, the source of meaning of the practice is in the value, courtesy, that it serves, and each participant in the social practice that serves the value must decide what it requires both for herself and for the community. Dworkin, *supra* note 3 at 63–64. Thus, a practice becomes meaningful insofar as it serves the value, not to the extent to which it corresponds to an announced rule.
[164] *Id.* at 3.

To briefly summarize his theory, Dworkin believed that good legal decisions require the satisfaction of two criteria: first, that they "fit" with the overall system of the law in the context of which they are decided and second that they put that system in the "best light" possible, or "justify" it according to principles like integrity, justice, and fairness.[169] In Dworkin's view, "the law is structured by a coherent set of principles about justice and fairness and procedural due process, and it asks [judges] to enforce these in the fresh cases that come before them."[170]

In practice, these two criteria require a multi-step process. A judge is presented with a case. First, she must come up with a set of theories for holdings that could resolve the case before her and then check those theories against the precedent cases that form the content of the law of the area at issue.[171] Theories that do not match most of the most important cases are thrown out.[172] Developing this set is the first part of fit, the requirement that the new decision accord with the old. But this is not simple statistical best fit like we might see in machine learning: for Dworkin, morality operates here too, and theories generated by the judge must both match the precedents and "state a principle of justice" that is connected to some "more general moral or political consideration."[173] In this way, the judge is already unlike an AI model simply trying to fit to all of the data that it is given. Next, the judge must expand the range of fit, looking to "the great network of political structures and decisions of [her] community" and aiming with the candidate theories to "justify[] the network as a whole."[174]

---

[165] *Id.* at 249–50 (discussing how an ideal judge should behave in a situation in which abstract justice and political fairness pull in different directions in resolving a case).
[166] *Id.* at 45–46.
[167] *Id.* at 44.
[168] *Id.* at 47–48.
[169] *Id.* at 239.
[170] *Id.* at 243.
[171] *Id.* at 240–42.
[172] *Id.* at 242.
[173] *Id.*
[174] *Id.* at 245.

Whatever candidate theories survive the fit step are subjected to the "best light" test, in which the judge decides which interpretation "shows the legal record to be the best it can be from the standpoint of substantive political morality."[175] This substantive political morality is made up of "abstract justice," apparently based on the judge's own sense of that, and "political fairness," the views of the political community of which the judge is a part.[176] Where these two components conflict, the judge must decide herself which to prefer, subject to an individualized higher-level sense of how the two interact in different legal contexts.[177] It is not clear whether a Dworkinian legal system staffed exclusively by Herculeses would be one in which the mechanisms of this balancing led to consistent applications of the law across contexts, but it seems likely that the ingredient of political fairness would lend it a degree of flexibility in its application. However, the Dworkinian structure of fit and best light and the idea that a higher-level sense of the balance between fairness and justice should structure adjudication lends Dworkinian interpretation a kind of "secondary rule" that gives it consistency. These structures operate as a set of meta-rules, creating a hierarchy of sources or process through which interpretation should happen.

As Dworkin summarizes his theory, fit "will provide a rough threshold requirement that an interpretation . . . . must meet if it is to be eligible at all."[178] Then, to decide among the surviving interpretations, the judge "ask[s] which shows the community's structure of institutions and decisions . . . . in a better light from the standpoint of political morality."[179] Importantly, the moral aspects influence how the words in the law should be interpreted, guiding the decisionmaker to a certain part of the space of meanings that could exist for any given word such that their ultimate decision at the level of language is shaped by morality. For Dworkin, one of the overarching goals of the law is integrity, which consists of the process of bringing the law into some kind of coherent picture that makes each of the elements of that picture the best that they can be.[180] The law is a system that is motivated by higher goals in service of the good of the political community, and the judge's role is to decide in ways that move toward the accomplishment of that objective in each case they adjudicate. Dworkin's vision is fundamentally a moral one, and it is through the application of high-level principles of political morality like justice and fairness[181] to the facts of specific cases that a system of law is justified and made legitimate.

### B. Analogies and incompletely theorized agreements

[175] *Id.* at 248. Because interpretations that do not have any relationship to political morality are eliminated at the fit step, all remaining interpretations can be evaluated according to whether they put the law in its best moral light.
[176] *Id.* at 249.
[177] *Id.* at 250.
[178] *Id.* at 255.
[179] *Id.* at 256.
[180] *Id.* at 225.
[181] In his later work, Dworkin argued for the unity of value, that these concepts are the same or at least mutually supporting, *see* RONALD DWORKIN, JUSTICE FOR HEDGEHOGS 1 (2011), but he makes a distinction in his work most directly relevant to our subject.

Contra Dworkin, Sunstein argues that general principles of abstract political morality provide a poor ground for legal reasoning. Instead, for Sunstein, the ground of the law is in concrete precedent, and though the moral views of individual people may inform how the law is made, the form of reasoning that lawyers and judges distinctively engage in is analogical reasoning[182]—which may lead the law away from what is morally legitimate[183]—not philosophy.[184] Both Sunstein and Dworkin agree that the law is composed of a set of fixed points, laws and cases that derive from our considered judgments about the world,[185] that form the basis for reasoning. Dworkin thinks that these fixed points are particularly important precedents that create a "fit" criterion,[186] or per Sunstein's description, "constrain the category of permissible general theories,"[187] but that the law must ultimately be interpreted in light of general moral ends expressed as principles. Sunstein is skeptical of the possibility of general theorizing altogether and argues that analogical reasoning, and thus the law, ends "at just the point when the relevant principles go beyond a low level of generality."[188]

The two authors' examples of legal reasoning provided in their works instructively underline the differences in their approaches. Dworkin insists that judges start with theories: "[Hercules, the ideal judge,] begins by setting out various candidates for the best interpretation of the precedent cases even before he reads them."[189] Only then should the judge see how the precedents fit to the theories that the judge has independently generated.[190] Sunstein also begins with an asserted proposition, which he calls "low-level,"[191] and then immediately makes sense of the proposition by analogizing its content to precedent.[192] Importantly, these approaches diverge not just on prioritization but on where they find the source of meaning for language. For Sunstein, a sentence does not make sense except insofar as we can place it in our mental context through analogization. The word "vehicle"[193] is just a sound until tied to a representation with particular features, and new stimuli are categorized as "vehicle" or "non-vehicle" based on the extent to which they resemble, or are analogizable to, the representation that we have.[194] In the law, precedents function as this kind of representation, providing categories that can make sense of the words in new statutes and cases.[195] For Dworkin, in contrast, the meaning of important legal words or practices must be understood with reference to values that underpin and justify them. As his example of courtesy shows, it is not sufficient to simply provide a set of examples of people being courteous to know what courtesy means because in a different context different (indeed contrary) practices could be understood to be courteous.[196] Instead, in Dworkin's view, disagreements about what words or practices to use emerge from real, theoretical disagreements about the deeper nature of the values of a particular community as it expresses itself. The words of a law should be read in light of the values of those who live in the community rather than in light of past interpretations of the law's constituent sentences' semantics. The meaning of a particular practice of courtesy, then, depends on the meaning of the principle itself and not on how it fits in with past practices. Hart, Dworkin, and Sunstein are all writing in the shadow of Wittgenstein, following his arguments that meaning emerges from the interaction of people in a community.[197] They diverge on where and how that meaning emerges.

---

[182] Sunstein, *supra* note 44 at 741–42.
[183] *Id.* at 759.
[184] In his article *On Analogical Reasoning*, Sunstein specifically calls out Dworkin's approach for being inconsistent with actual legal practice, in which lawyers do not engage in "general moral theorizing" but seek to address cases by reasoning analogically. *Id.* at 784–85.
[185] *Id.* at 751.
[186] Dworkin, *supra* note 3 at 238–39.
[187] Sunstein, *supra* note 44 at 753 (citing Brown v. Board of Education, 347 U.S. 483, Roe v. Wade, 410 U.S. 113, and affirmative action as examples). Obviously, *Roe* has turned out not to be a fixed point that constrains subsequent decisions.
[188] *Id.* at 754. Here, Sunstein seems implicitly to be asserting a version of the separability thesis. *See generally*, H.L.A. Hart, *Positivism and the Separation of Law and Morals*, 71 HARV. L. REV. 593 (1957).
[189] Dworkin, *supra* note 3 at 240.
[190] *Id.*
[191] Sunstein, *supra* note 58 at 759.
[192] *Id.* at 760.
[193] *See* Hart, *supra* note 1 at 123; for a more complete treatment of this example, see H.L.A. Hart, *Positivism and the Separation of Law and Morals*, 71 HARV. L. REV. 593, 607 (1958).

Beyond the descriptive arguments, for Sunstein there is a normative value to analogical reasoning as much as it is also to him a more accurate descriptive picture of how lawyers proceed. Sunstein argues that coming to agreement on general theories is nearly impossible and largely inconsistent with the goal of creating a pluralistic society in which people of different perspectives can live together.[198] As Sunstein writes, analogical reasoning consists of lawyers "develop[ing] low-level principles to account for particular judgments, and apply[ing] those low-level principles to new cases in which there is as yet no judgment at all."[199] These low-level principles then become concretized over time, taking on the status of precedent as they are accepted by society.[200] The law is the mechanism of analogical reasoning on which Sunstein focuses, but it seems reasonable that an even more broadly democratic or pluralistic mechanism of deliberation could be developed that extended these benefits. People reason analogically all the time, and often ues this reasoning process to come to agreement on practical questions of daily life. The key benefit of analogical reasoning is that, because of its proximity to life in the world, it can specify things in the world more clearly than through general rules. Sunstein's theory concentrates on concrete particulars and the process by which those particulars can be made to stand in relation to each other, to cohere.[201] Analogical reasoning is that process.[202]

---

[194] Debates in psychology over whether these representations are defined by statements of collections of relevant concepts or simply sets of examples of those concepts are ongoing, and seem to closely track the differences between Hart, who thinks of cases as subordinate to rules, and Sunstein, who puts cases first. *See* Gregory L. Murphy, *Is there an exemplar theory of concepts?*, 23 PSYCHONOMIC. BULL. REV. 1035, 1035 (2016).
[195] Hart argued similarly that sentences of law could be understood as having core, relatively uncontested meanings in the form of similar examples repeated across contexts. Hart, *supra*, note 1 at 123–24.
[196] *See* Dworkin, *supra* note 3 at 47–49.
[197] And indeed, they all cite him in the works being analyzed here. *See, e.g.*, Hart, *supra* note 1 at 297 n. 125 (citing Wittgenstein's *Philosophical Investigations*); Dworkin, *supra* note 3 at 63 (same); Sunstein, *supra* note 58 at 753 n. 43 (same).
[198] Sunstein, *supra* note 59 at 1735.
[199] Sunstein, *supra* note 58 at 758.
[200] *Id.* at 771.

Overlaying, and resulting from, the process of analogical reasoning is a set of "incompletely theorized agreements," that participants in the legal reasoning process come to about the outcomes of cases without ever having to agree on the deeper reasons behind particular decisions.[203] As Sunstein writes, these participants "agree on the result and on relatively narrow or low-level explanations for it. They need not agree on fundamental principle. They do not offer larger or more abstract explanations than are necessary to decide the case."[204] This specificity is one of the great strengths of legal reasoning, in Sunstein's view. Because legal reasoning does not rely on coming to agreements about higher principles like morality and the greater good, people with different values can still accept specific outcomes of the law even if they would have come to those outcomes by different paths. Thus, analogy provides the mechanism to reach incompletely theorized agreements, which then provide the content of the law. A Sunsteinian judge also does something like fit, but rather than reaching for a principle like justice to evaluate the candidate theories that survive that process, she remains at a lower level, deciding this case like others have been decided.

Two immediate objections to Sunstein's arguments rear their head. First, while it may be true that analogical reasoning is a particular characteristic of legal argumentation, it is not clear what analogies are made of. Sunstein anticipates this critique and explains it in depth, writing that because "[e]verything is similar in infinite ways to everything else, and also different from everything else in the same number of ways," "one needs a theory of relevant similarities and differences."[205] Because analogical reasoning cannot provide this from itself, "[i]t is thus dependent on an apparatus that it is unable to produce . . . . At the very least one needs a set of criteria to engage in analogical reasoning. Otherwise one has no idea what is analogous to what."[206] But he also has answers. First, Sunstein argues that general theories are also unsatisfactory, too inflexible to provide a full account of how to balance the various desired goods that may be traded off through different ultimate decisions.[207] More deeply, however, Sunstein believes that low-level convictions about individual cases reached through analogical reasoning deserve priority over general principles, and that "there may be no criteria for truth in law except for our considered judgments about particular cases, once those judgments have been made to cohere with each other."[208] Further, "coherence in law might then be defined as consistency among particular judgments and low-level principles,"[209] which are tested over time with reference to other low-level and high-level principles.[210] On this view, analogical principles take a kind of canonical status over time as they are enmeshed in a framework of other principles and fixed points, taking meaning from their presence within and among those existing structures. The law is just these principles, and the content of political morality emerges from the network of them. Low-level principles emerge from the network of precedents itself, bootstrapping a system into existence. Because people already know what words basically mean and can apparently reason analogically without having been trained to do so, they can build such a system. In later work, Sunstein concedes that this answer does not fully resolve the problem of how to decide among infinite possible analogies, but he does believe that analogy remains the essential engine of the law even if some theory is necessary.[211]

---

[201] *Id.* at 775–76
[202] *Id.* at 775.
[203] Sunstein, *supra* note 59 at 1735–36.
[204] *Id.*
[205] Sunstein, *supra* note 58 at 774.
[206] *Id*. It is instructive to note that this feature of analogical reasoning was also pointed out by Hart, who wrote that judges can never exhaust the various dimensions of distinction that might exist among precedents so as to come to a fully specified new decision. Hart, *supra* note 1 at 134-35.
[207] *Id.* at 776.

Second, a critic might argue that without a general principle to determine what kind of analogy is a good one, it is impossible to determine whether a given analogy is just or should be used. In other words, in precisely the hard cases where the statement at issue is most ambiguous or dissimilar to precedent, analogical reasoning fails as a guide and the judge can pick whatever analogy they want. In his more Dworkinian moods, Sunstein concedes that it is necessary to find a high-level way of determining what kinds of analogies are the right ones to apply to new situations.[212] But he has two major responses to this kind of critique. First, Sunstein argues that this problem is a feature of general principle-based systems, and worse there. He gives the example of Holmes's infamous opinion in *Buck v. Bell*,[213] the involuntary sterilization case, arguing that though Holmes did use analogy in support of finding that involuntary sterilizations could be mandated, he actually was reasoning from "a principle of a high level of generality," the general public welfare, that was "not evaluated by reference to low- or intermediate-level principles that may also account for the analogous cases."[214] For Sunstein, Holmes's failure was that he was not reasoning in a constrained way, considering all relevant precedents, but rather reaching for a particular outcome because he preferred it on principle and then justifying it *post hoc* by means of analogical reasoning. Judges following general principles can operate in an unconstrained way because they can fashion arguments for the application of those principles in whatever way they want.[215] Analogical reasoning, if done correctly, helps constrain and guide the grounds of decision.

Sunstein's second major response is that low-level principles should be preferred to general principles because low-level convictions about specific cases "deserve priority in thinking about good outcomes in law."[216] Because, as discussed above, coherence among considered judgments about particular cases is the only criterion for truth in law,[217] the law takes its force and meaning from the accumulation of principles and precedents that structure it. For Sunstein, it is the low-level judgments that give meaning and content to legal reasoning and that provide the grounds of the law, not some reference to transcendent values.

---

[208] *Id.* at 777.
[209] *Id.*
[210] *Id.* at 778.
[211] *See* Sunstein, *supra* note 60 at 32.
[212] Sunstein, *supra* note 60 at 31–32.
[213] 274 U.S. 200 (1927).
[214] Sunstein, *supra* note 58 at 757.
[215] *Id.*

In short, Sunstein and Dworkin provide opposing theories of how meaning emerges in the law, though as we shall see their approaches are actually in many ways mutually supporting because theory requires application to give it meaning and analogy requires theory to give it structure.[218] Responsibly choosing a given part of the sets of meanings that a legal statement can plausibly have requires referring not just to the likelihood that a particular word was used in a particular way, but reading the language in light of the values and agreements, at whatever level, that actors in the community share. As such, a mixed approach or simultaneous seems to be best, meaning bootstrapping itself into existence from the background of existing language and the world,[219] principles and examples reinforcing each other through content and application.

## *III. Alignment as jurisprudence*

The isomorphism of alignment and the law means that the legal theories outlined above can aid alignment researchers in identifying problems of alignment and the solutions to them. In particular, legal theory offers solutions to the problems of representation and specification that alignment techniques currently face. Existing alignment techniques are either unrepresentative or majoritarian, using principles and values either thought up by small groups in AI labs or representing the simple majority views of small polled populations and applying them in ways that exclude alternative perspectives. In either case, the diverse views and values represented in society are not represented in alignment, and as AI models take on increasing power and responsibilities, if minority views remain unrepresented then they may be crushed by the operations of AIs that operate in ways contrary to them. Even if a path to a more pluralist form of alignment is found, existing techniques are relatively unable to specify the content of the principles or policies that are put into models. Currently, these principles are expressed as general statements of rules, but the ambiguity and contextuality of natural language makes the application of these rules uncertain and difficult for researchers to guide with any degree of granularity. Simple statements that models should be "helpful" or "fair" leave the interpretation of those principles up to the models based on their underlying understanding of the words in the principle and risk having the model interpret the principles in ways contrary to the intent of aligners or what is best for society. The law has solutions to these problems. Each of the legal theorists discussed in the previous Part sought answers to the specification problem, arguing that different sources of meaning could be used to specify the content of rules in their practical applications, and that through the use of precedent these applications could accumulate to create a network of past points into which new interpretations can be fit. Each of them also emphasized the importance of providing ways for the political community subject to the decisions of judges to be reflected in the reasoning process of those judges, ensuring that their decisions were representative. Because the law and AI share the same shape, these theories can help answer how to resolve these key questions in alignment.

[216] *Id.* at 777.
[217] *Id.*
[218] As Sunstein ultimately concluded. Sunstein, *supra* note 60 at 31–32.
[219] As a large language model uses the underlying background of its training on language to generate a set of representations of the world before scoping in on their relationships to generate particular outputs.

As discussed,[220] in both the law and in AI a set of prior decisions provides data used by the judge or model to respond to some new input. In the legal system, the decisions, statutes, constitutions, and other documents that form the basis of legal reasoning are both substantive precedents that directly shape how the new case will be decided, in part by giving meaning to the underlying sources of law that the new decision will draw on, and are also examples of the kinds of reasoning that the legal system is built to do. Judges making new decisions thus reason in both the content and form of the system, extending it in their new opinion. The models of legal reasoning put forward by Hart, Dworkin, and Sunstein converge here, each of them emphasizing the role of precedent in binding the next decision into some relation, whether principled or analogical, with the former ones. For Hart and Sunstein, meaning arises from social practices and past interpretations, while for Dworkin background values are necessary to give legitimate meaning to applications or interpretations of the law. Foundation models get their examples and patterns of reasoning from their training data, a form of precedent in the use of language, and from particular examples made salient by finetuning. A huge amount of training data goes into each model and it learns the patterns that structure the data, from the likelihood of some word coming after another word or set of words[221] to working across modes of text, images, and sounds.[222] The model applies these patterns to each new input that comes before it like a judge extrapolating from precedent to decide a new case based on the facts and holdings of those cases.[223] Then, researchers seek to give the models substantive guidelines and constraints on what extrapolations they choose through the use of alignment techniques like those discussed above,[224] for example by pointing out specific cases to the AI and making them more salient in its reasoning process through the use of finetuning techniques like RLHF and its descendants, as well as through inference scaling and similar innovations.[225] In a sense, finetuning points the model to the part of its underlying distribution or map of meaning that the researchers want the model to draw on. The training data provides a linguistic and conceptual background for reasoning and the finetuning process functions like the set of precedents on point, focusing the AI on a particular part of the conceptual space that is most relevant to its decision. This process is like the processes of interpretation through value or example that Dworkin and Sunstein put forward, using precedents and patterns of past interpretation to inform the meaning of ambiguous rules in new cases.

---

[220] *See supra* Section I Part A & Section II.
[221] *See* Wolfram, *supra* note 7 at 2.
[222] *ChatGPT can now see, hear, and speak*, OPENAI (Sep. 25, 2023), https://openai.com/blog/chatgpt-can-now-see-hear-and-speak.
[223] Wolfram, *supra* note 7 at 1.
[224] *See supra* Part I.
[225] For example, one regularly suggested tip for using consumer models like ChatGPT is to "few-shot" it, to give it a few examples of the kind of reasoning that the user is asking for, in order to get it in the mode of that kind of reasoning before the user asks the question that she is seeking to get an answer to. *See* Tom B. Brown et al., *Language Models are Few-Shot Learners*, ARXIV 4–5 (Jul. 22, 2020), https://arxiv.org/pdf/2005.14165.pdf

These legal theories can help alignment with its problems of pluralism and specification. Below, I match Dworkinian interpretivism to Constitutional AI and Sunstein's analogical reasoning to CBR, illustrating specific insights that can be drawn across from the law into alignment. Each of these concepts is far deeper and more complex than I have been able to cover in this short space but, in short, the problems of meaning specification that both alignment approaches are facing could be addressed by using legal concepts. Constitutional AI should try to incorporate a hierarchy of principles that are used to provide guides for the application of rules in specific contexts and generate those rules by reference to deliberation about cases, as Dworkin laid out in his framework of best light and fit.[226] Creating a structure in which certain principles are elevated to a higher or legitimately "constitutional" level[227] and then having those principles inform the application of the lower principles would provide a guide for how to resolve questions of ambiguity in their application in the same way that higher values inform Dworkinian interpretation. Combining such a structure with a system of "fit" or precedent that allowed for specification of the meaning of the principles by filling in their open texture through examples of approved applications in edge cases, as Hart and Sunstein argued for, would deepen the extent to which the principles could be predictably applied in varying cases. Thus, for example, master concepts like justice, equality, or fairness[228] could be used to inform how subsidiary principles are applied in concrete cases and each application of those principles would ramify back into the system, providing greater definition. Because public reasoning about concrete cases would represent a core part of how the meaning of the values emerged, this system would be reasonably democratic.

The CBR approach should similarly be modified to incorporate direct public reasoning about alignment and what approaches to it should be taken. Applying Sunstein's insights, CBR should take a set of applications of AI in the world and then have people deliberate about whether those applications are the best ones for AI to be doing. Their decisions and their reasons for them could then be brought back into the system as fixed points that could be used to guide the actions

---

(demonstrating that large language model performance can be improved by few-shot learning). For a discussion of inference scaling, see Yangzhen Wu et al., *Inference Scaling Laws: An Empirical Analysis of Compute-Optimal Inference for Problem-Solving with Language Models*, ARXIV (Mar. 3, 2025), https://arxiv.org/abs/2408.00724.

[226] Dworkin, *supra* note 3 at 256.

[227] As opposed to the more "statutory" level of current Constitutional AI principles that all exist at the same level of importance."

[228] Insofar as they are distinct. *See generally* RONALD DWORKIN, JUSTICE FOR HEDGEHOGS (2011); JOHN RAWLS, A THEORY OF JUSTICE (1971).

of models in the future. Because LLMs can generate large quantities of varying text, they could be used to generate diverse cases for humans to deliberate about, and as the deliberation proceeded it could be used to influence what kinds of cases the models generated in future. Thus, once a fixed point about some AI action emerged, future cases that are in varying ways "like" that action could be presented to the users. However, this "likeness" should emerge simply from having the generative model iterate across the possibility space rather than having experts define it. Over time, some sense of the relevant similarities and differences among these cases might emerge that could form the basis of further extrapolation.

A combined approach would involve using CBR to specify the meaning of the principles that are used in Constitutional AI. This approach might actually be the most effective and easiest to introduce, as the existing Collective Constitutional AI deliberation platform could simply be modified to produce sample cases based on the principles that users select or write and then present those cases to the user for decision and analysis. Thus, users would be able to reason at a high level about principles and then specify what they mean by the principles through selecting applications that align with their idea about the meaning of cases. Cases that implicate multiple principles could be presented to other users who picked different of those principles, and where many users coming from different principles selected the same outcome, there would be an incompletely theorized agreement. Combining Dworkin and Sunstein in this way would allow for both more democratic and more effective specification of meaning in the alignment system and would likely not represent an overwhelming engineering burden. Empirical experiments among these applications are necessary to determine which of them would actually be successful in improving alignment, but they each represent a useful potential path forward.

### A. Interpretivism and Constitutional AI

Constitutional AI resembles Dworkin's theory of the law, and elements of Dworkin's approach can help resolve some of the limitations of Constitutional AI's current form. The conceptual similarity between the two approaches relatively clear, though there are some differences in how they operate and any cross-application must be stylized. As discussed above, Dworkin's "fit" maps onto the pretraining stage of making an AI model. In each version, the judge or AI consumes a set of training data and figures out the conceptual patterns underpinning that data. These patterns are then used to ground future decisions or completions. There are some differences between Dworkinian and statistical fit. Most importantly, Dworkin does not simply allow for every possible theory that explains most or the most important precedents to survive the fit stage.[229] Instead, he eliminates candidate interpretations that do not announce a plausible "principle of justice" and so fail to connect the caselaw to a "more general moral or political consideration."[230] Under the "shovel everything in" approach to training LLMs, in contrast, pretty much all of the data that is available in the world is put into the model for it to fit to,[231] and the models do not seem to apply any moral or political considerations to evaluating their own training data at the pretraining stage.[232] While AI labs do seem to exclude some data on the basis that it contains objectionable content,[233] that is different from the internal application of principles by a model choosing how to understand the patterns that structure its pretraining data. Models are incapable of such operations on their own, especially during pretraining, and there are interesting questions about what would happen if all sexist data, for example, were excluded from the pretraining process of a model, if such a thing were possible.[234] One potential Dworkinian improvement on the Constitutional AI approach would be to train a model to recognize racism or some other objectionable type of content and then have it go over training data, modifying it so as not to contain those patterns, and then train a new foundation model on the modified data. Thus, the modifying model could act as a kind of fit-level imposer of principles that might remove those harmful patterns from the newly trained model, though it might be better simply to do alignment later, so that, knowing evil, the model could identify good.

[229] Dworkin, *supra* note 3 at 242 (eliminating the second of six candidate theories for resolving a sample case because it does not announce a principle of justice).

Next, Dworkin's justification stage maps onto the finetuning alignment stage of training an AI via Constitutional AI. In the justification stage, moral and political principles like justice and fairness are applied to the judge's remaining theories of the cases, and the judge selects the theory that puts the whole system of the law in the best light according to those moral principles.[235] The rule at issue is interpreted in the way that best supports the relevant principle. Similarly, Constitutional AI seeks to use its Constitutional principles to get the model to select the candidate outputs that most closely align with the principles that the model creators have chosen and introduced into it.[236] In an important sense, the designers of Constitutional AI actually acted like Dworkinians in choosing their original set of principles to use to govern the models. Their justification for their reliance on the Universal Declaration of Human Rights and on other documents that they believe represent some kind of social consensus on morality[237] resembles Dworkin's arguments for relying on the political morality of the community as a whole, itself a kind of agent,[238] which generates principles through people living and deliberating together.[239]

---

[230] *Id.*
[231] *See* Deepa Seetharaman, *For Data-Guzzling AI Companies, the Internet Is Too Small*, WALL ST. J. (Apr. 1, 2024), https://www.wsj.com/tech/ai/ai-training-data-synthetic-openai-anthropic-9230f8d8.
[232] In fact, LLMs seem to replicate human cognitive biases and common errors of reasoning, suggesting that they are just finding whatever patterns, however erroneous, exist in their training data. Erik Jones & Jacob Steinhardt, *Capturing Failures of Large Language Models via Human Cognitive Biases*, NEURIPS 2022 8–10.
[233] *See* Matthew Hudson, *Robo-writers: the rise and risks of language-generating AI*, NATURE (Mar. 3, 2021).
[234] It's not clear, for example, whether a model could then recognize sexism such that it could even acknowledge that it exists in the world when prompted later on. *Id.* (quoting Amanda Askell making this point).
[235] Dworkin, *supra* note 3 at 245.
[236] Bai et al., *supra* note 22 at 1–2.
[237] *Claude's Constitution*, *supra* note 11 at n. 2.
[238] Dworkin, *supra* note 3 at 187–88.
[239] *Id*. at 189–90.

Both approaches rely on committing to relatively abstract sets of high-level values that emerge and become concrete through a process of collective reasoning. Dworkin writes of justice and fairness, and each of these is given meaning through the interpretive practice of the community.[240] The Constitutional AI researchers, especially in their work on using general principles, similarly do not prescribe or proscribe particular kinds of behavior but rather provide abstract values that take on content from their meaning in the training data that underlies the models, effectively from the general uses of the words that make up the principles on the internet, which is some kind of community.[241]

### *i. Pluralistic values and meta-principles*

The first major problem facing Constitutional AI that interpretivism can help address is that it is not representative of the richness of values of the society in which and on which the models will be operating. The collective constitution approach explored by Anthropic researchers,[242] while a promising step forward, does not solve this problem because it is unable to balance democracy with protections for the rights of minorities. Foundation models are, in fundamental ways, majoritarian machines. The weights in their neural networks, which provide the mechanism by which they process inputs and generate outputs, are majoritarian, in that whatever content is in the majority in their training data about something ends up being how they process that thing.[243] So, for example, if the majority of the training data ascribes a particular negative set of characteristics to some political party, then the model will generate outputs characterizing the party in that way, unless it model is finetuned to prevent that. Thus, minority viewpoints and perspectives about controversial issues are unlikely to be what is output by the model unless society decides to protect those outputs. But some interventions will only reproduce the majoritarian bias. For example, the collective constitution created by Anthropic in collaboration with the public was the result of a majoritarian process.[244] The principles included in the collective constitution were those selected by a majority view of the voters on each statement, while minority views were left aside.[245]

Protections for minority rights are an essential element of a pluralistic democratic society, and one of the foremost roles of constitutional law is to ensure that minorities are protected.[246] Dworkin was aware of this role, and he believed that judges should be less willing to listen to the wishes of the majority of their community in matters of the protection of minority rights.[247] In Dworkin's view, these situations are ones in which justice and fairness, understood respectively as an abstract principle and the sense of political morality of the majority of the community, are in tension with each other, and that here judges should weight justice more heavily.[248]

[240] *Id.* at 247–48.
[241] *See* Kundu et al., *supra* note 112 at 12–13, 24.
[242] *See supra* Part I Section B Subsection i.
[243] Sorensen et al., *Value Kaleidoscope*, *supra* note 36 at 19938.
[244] Ganguli et al., *supra* note 106.
[245] *Id.* (note that the values of Group B were not included in the collective constitution).
[246] West Virginia State Bd. of Educ. v. Barnette, 319 U.S. 624, 638 (1943) ("The very purpose of a Bill of Rights was to withdraw certain subjects from the vicissitudes of political controversy, to place them beyond the reach of majorities").

To give more content to Dworkin's argument here, it is useful to turn to the function of foundation models as contextual reasoning machines. Dworkin's claim that judges should weight some elements of morality more heavily in certain cases is an argument that some kinds of adjudications are different from other kinds of adjudications, on the basis of the claims at issue and the implications of the case for society. That argument is also the basis of constitutional law, which takes on a status superior to the dictates of both legislatures, in the form of statutes, and judges, in the form of common law decisions. Changing constitutional law requires almost the whole of society to speak with one voice,[249] with the goal of ensuring that any such fundamental change be one of which many minority groups approve.[250] But knowing when a question is a constitutional one or one implicating basic rights is a matter of understanding the context of the claim that is being presented. Dworkin's judge must look to the content of the case in front of her and to the implications that a particular decision in that case will have on society—in some sense, "best light" is defined by considering how a given decision will change the context of political morality by adding another element to the picture.

The upshot of all this for making AI more representative and pluralistic is that models need to be given meta-principles telling them how to think about balancing the values they are charged with enacting. Dworkin wrote that when "[justice and fairness] conflict," judges will have to decide which to pick in order to "show the community's record in the best light."[251] To do so, they will have to have "higher-order principles" to evaluate the clash, for example the belief that while "political decisions should mainly respect majority opinion . . . . this requirement relaxes or even disappears when serious constitutional rights are in question."[252] Truly constitutional AI must provide a superset of principles that can be used to provide a mechanism of decision in cases of tradeoffs among values in the same way that Dworkin's ideal judge balances higher values in adjudicating cases. Foundation models' ability to reason contextually will allow them to apply those meta-principles effectively in different cases, deciding whether some circumstance requires the invocation of the higher law with which the model is imbued. These meta-principles should include protections for minority rights and a kind of embrace of pluralism, thus allowing for a model to serve the necessary role of guarantor of non-majority perspectives and to evaluate when and how to generate outputs protecting them. It is not clear how meta-principles should be put into models using current approaches. Simply including meta-principles in the sets of principles that models are trained on is one potential approach, especially if they are set apart in some way and their special status is indicated. However, empirical research investigating how to strengthen the extent to which these principles are considered first would be useful.

---

[247] Dworkin, *supra* note 3 at 257.
[248] *Id.* at 256–58.
[249] This is the picture of changing constitutional law by amendment. *See* Office of the Federal Register (OFR), *Constitutional Amendment Process*, NATIONAL ARCHIVES (Aug. 15, 2016), https://www.archives.gov/federal-register/constitution. Of course, constitutional law is also (or, these days, perhaps only) made by the Supreme Court, though they would claim that they are not changing the constitution but merely applying what it already says with the voice of the people. *The Court and Constitutional Interpretation*, SUPREME COURT OF THE UNITED STATES (n.d.), https://www.supremecourt.gov/about/constitutional.aspx. Especially since the establishment of the congruence and proportionality test, limiting Congress's power to enforce constitutional rights, *see* City of Boerne v. Flores, 521 U.S. 507, 530, 532 (1997), the Court has had effectively sole control over constitutional lawmaking.
[250] Erwin Chemerinsky, *Amending the Constitution*, 96 MICH. L. REV. 1561, 1561 (1998).
[251] Dworkin, *supra* note 3 at 256.
[252] *Id.* at 257. In some sense, these meta-principles resemble Hart's "secondary rules." *See* Hart, *supra* note 1 at 93-99.

*ii. Substance and scaling*

The other major problem facing the Constitutional AI approach is that it does not address the vagueness and indeterminacy of the meaning of the principles imbued in the models, making scaling alignment difficult and risky. Constitutional AI needs to have some theory of how the principles work; the current approach of simply telling AI to be good or fair or not to discriminate on the basis of race or gender has had remarkably successful results given how simple it is, but it does not provide a path toward predicting what a model will do in some new context besides just some approximation based on a general sense of what words mean. If you tell a model to be "just" in its outputs, what have you actually told it to do? The researchers behind Constitutional AI are aware of this problem.[253] Writing about their experiment in using a principle expressed at a very general level, they say "[t]he 'good for humanity' approach has a potentially huge problem–it simply leaves the interpretation of the [good for humanity] idea to AI systems themselves. This interpretation will necessarily be culture-bound, and is likely to vary by language and era. For any given AI system, it presumably is determined in some complex way by the distribution of pretraining data."[254] The benefits of transparency gained from constitutional AI[255] are thus limited because they only go so deep. Particularly as models are given increasingly important tasks and begin making decisions in the world, this vagueness about the meaning of the principles risks making the extent to which they are actually aligned in any new scenario a real question.

Can Dworkin help here too? His interpretivism seemingly suffers from the same problem as the one just described, as it is apparently impossible to determine from the outside what a judge will do in a given case because we cannot understand what they believe justice and fairness mean.[256] Dworkin conceded this difficulty, writing that each judge will come to rely on "a fairly individualized working conception of law on which he will rely," rendering judgments "a matter of feel or instinct rather than analysis."[257] He concluded that it is only possible to represent this individualized conception as a set of "rules of thumb" that the judge should generally apply, subject to the requirements of changing circumstances.[258] Dworkin seems to have believed that this pattern of human conduct, or perhaps limit on human perspicacity or on language itself,[259] is just the substance of what the law is, each author making her own idiosyncratic contribution to the great chain novel[260] which emerges from the conduct of human life.[261] While a useful metaphor, such a concession also entails giving up on ever effectively specifying how to decide in the future.

---

[253] Kundu et al., *supra* note 112 at 24–25.
[254] *Id. Cf.* Hart, *supra* note 1 at 123 ("Canons of 'interpretation' cannot eliminate, though they can diminish, these uncertainties; for these canons are themselves general rules for the use of language, and make use of general terms which themselves require inter-pretation. They cannot, any more than other rules, provide for their own interpretation.").
[255] As compared to RLHF, in which even the finetuning of the AI is impossible to really understand because it is based on humans giving feedback rather than on clear principles. Bai et al., *supra* note 22 at 3–4.

However, combining Constitutional AI with example-based specification based on principles might provide a path forward here. As Dworkin argued, the set of practices that might serve a particular value can change over time, even as that value remains consistent itself.[262] Thus, an observer from outside a society might see the practices that relate to a given value diverge, or even become contradictory,[263] over time though those within the society acknowledge that each practice in fact serves the underlying value. Such a condition seems to make specification through example difficult. However, there must by definition be some through-line that unites these practices and gives them a relation to the underlying value.[264] For Dworkin, the through-line is the moral value itself. Each new social practice in a given set must reflect in some way the underlying value that defines that set if it is to be included in it, and new practices are to be interpreted in light of the value. A diversity of examples, both of the scope of practices that relate to a given value and of the changes in those practices over time, is actually likely the best way for a classifier to determine what kinds of future practices and decisions could correspond to that value. Thus, Constitutional AI should be supplemented with a set of human decisions about different cases related to its principles that illustrate them drawn from a diverse set of contexts and cultural backgrounds. The more diverse these inputs are, the better, as the model could then more effectively learn what the meanings of the principles are by seeing what elements of the training data are deemed to correspond with the values of that they are intended to illustrate. The Polis system that was used to facilitate the creation of the collective constitution[265] could probably be modified to include sets of examples generated by models with reference to the principles. Then, the participants in deliberation could reason not just about what principles or rules they prefer but actually how those principles should be applied in specific situations. The process of deliberation and the outputs generated would provide examples of application that better specify the principles and also examples of reasoning about rules and morality that could guide extrapolations of the principles in the future. The inclusion of examples featuring contradicting principles would also go some ways toward developing a more truly constitutional approach to alignment by allowing users to teach the model that certain principles or values are more important than others, making them a kind of higher law that could help guide the model through conflicts. Whereas human judges, in Dworkin's view, must rely on individualized "rules of thumb"[266] to resolve these kinds of values conflicts, AIs could rely much more closely on how the society that they are a part of thinks about the interactions between values, allowing for a much more effective and representative form of alignment that could be scaled to new fact situations while grounded in human values.

---

[256] The meaning of a word like justice, for example, has been the subject of philosophical debate since at least the Ancient Greeks. *See generally*, PLATO, THE REPUBLIC (B. Jowett trans., 1998).
[257] Dworkin, *supra* note 3 at 256.
[258] Dworkin uses the term "rules of thumb" twice when describing a judge's system, first as the highest level of precision to which an analyst of the judge can aspire to, *id.*, and second as all that a judge should think of her principles as. *Id.* at 257–58.
[259] Hart agreed that this might be a fundamental limit of language. Hart, *supra* note 1 at 128.
[260] *See id.* at 228-38.
[261] *Id.*
[262] Dworkin, *supra* note 3 at 70–71. *Cf.* Lawrence Lessig, *Understanding Changed Readings: Fidelity and Theory*, 47 STAN. L. REV. 395, 402–07 (1995) (arguing that changed readings of constitutional text do not necessarily imply that the underlying meaning of the texts have changed).
[263] For example, some might see taking off a head covering during prayers to be a requirement of piety while others might believe that piety requires wearing such a head covering.
[264] Dworkin, *supra* note 3 at 69.

For Dworkin, law is both a social and a theoretical practice, rooted in a community's sense of what the world should be like but also aligned to higher values that legitimize the practice of that community. As AI systems grow in intelligence, they may take on conceptions of abstract values like "the good" that guide the way that they behave across contexts, even when those values are not expressly invoked.[267] These higher values might, as they do for Hercules, provide a baseline of meaning that AI systems can use to resolve cases of ambiguity and better serve the higher goods of the political community into which they are invoked. Here, values are not just invoked in the Constitutional AI sense of guidance but interact more fundamentally with the process of linguistic interpretation of rules, deepening the extent to which alignment to a broader society is occurring.

### B. Grounding Case-Based Alignment

Alignment by case-based reasoning (CBR) seeks to avoid the problem of the vagueness of general principles expressed in natural language by working from the ground up, using cases as fixed points that illustrate the meaning of rules and words by examples of their application.[268] In a fully developed CBR alignment approach, models would be given examples of different situations and their outcomes and then could identify the underlying principles that structure the decisions in those scenarios and extend them to new cases. But CBR still faces a problem like that facing Constitutional AI: The current version of CBR does not provide a way of identifying the content or structure of the extrapolations that will be made from the case repository to resolve the instant case. As Sunstein argued, any case is like, and unlike, any other case in an infinite number of ways, and the key is figuring out what dimensions of likeness and unlikeness are important.[269]

---

[265] *See* Ganguli et al., *supra* note 114.
[266] Dworkin, *supra* note 3 at 257–58.
[267] The extent to which Anthropic's models seem to be able to generalize and apply a concept like "good for humanity" supports such a claim, *see* Kundu et al., *supra* note 112 at 24–25, as does recent evidence that models have relatively robust conceptions of "evil" such that training them to be more malignant in one context makes them exhibit associated traits more generally. *See* Jan Bentley et al., *Emergent Misalignment: Narrow finetuning can produce broadly misaligned LLMs*, ARXIV (Mar. 5, 2025), https://arxiv.org/abs/2502.17424.
[268] *See* Levi, *supra* note 2 at 501–502.

The current CBR approach does not provide a technique for identifying what dimensions of analogy and distinction will be relevant for a model other than those selected by experts, which undermines the pluralism that the approach otherwise allows and is also less effective at finding the full set of possible meanings than an approach that is not directed by specific humans that is shaped by their biases. Given these limitations, it is impossible to fully predict how a model will extrapolate from its case repository to resolve a new case, reducing the extent to which CBR can be useful in constraining models to act in ways that are consistent with what the aligners want when they are trying to specify values and decisions ahead of time. Sunstein's work on analogical reasoning and incompletely theorized agreements[270] could help improve CBR by providing a framework for discussing how cases in the case repository fed to the AI relate to each other and what kinds of principles of analogy and distinction should be used in extrapolations. The law provides a rich set of examples of extrapolations from relatively settled precedents, and formal work specifying the content of analogical principles and the features of similarity and difference that they rely on in different cases could provide a ground for improving CBR and making it more effective and representative.

To review, in the existing CBR approach, a relatively small set of seed cases is turned into a full repository through the use of a large language model that iterates on the seed cases according to some set of dimensions of concern identified by experts in the domain of reasoning at issue.[271] Thus, for example, an expert might identify the age of users as relevant to a case about whether to put protections in place on certain internet services,[272] and then a generative model creates a series of iterations on the original case changing the ages of the users; these iterations are used to test what people think should be the rules in each of the different situations.[273] Then the underlying model can be trained on the full repository and its outcomes and extrapolate in new situations based on the original patterns of reasoning across circumstances.[274] Ideally, this model will learn how people think about the importance of age when addressing problems relating to the internet (or whatever topic is at issue) and then be able to apply that kind of consensus reasoning in the new contexts. In particular, it might learn that people think that minors should be treated differently from adults with regard to internet age protections and be able to generalize that to different contexts. In the language of Sunstein, each case generated from the repository is exactly analogous to the original seed case except on one dimension, that of age. Then, when the model sees where humans have decided some almost-perfectly analogous cases are actually distinct from the seed case, for example when decisions diverge because the minor becomes an adult, it will ideally realize that that particular age has become an important distinguishing dimension and figure out how to extrapolate from that distinction. The idea that provision of internet services to minors should be governed differently than provision of such services to adults is a kind of low-level principle that people might agree on.[275] They might not agree on why minors should be treated differently, or they might. But there is agreement on the outcome, forming an incompletely theorized agreement.[276]

---

[269] Sunstein, *supra* note 44 at 774.
[270] *See supra* Part II Section B.
[271] For a more complete discussion, *see* Part I Section C.
[272] Chen et al., *supra* note 82.
[273] *Id.*
[274] *Id.*

However, this approach does not help resolve the problem of figuring out what kinds of extrapolations are likely to occur given different seed and repository sets and does not allow for public reasoning about those modes of extrapolations. The current approach to CBR relies on iterating across one particular dimension of difference, like age, but does not provide a principle along which the kind of reasoning is done other than that it has been chosen by experts. In other words, it only iterates based on what some group already thinks the important dimension is rather than allowing the model to explore what a majority of humans might consider it to be. Furthermore, the more examples of different ages that are needed to illustrate a distinction to a model, the more difficult and intensive the process of specification will need to be up front, trading off with the efficiency of specification by example. Finally, the current approach does not allow for the full pluralistic exploration of the different ways in which a principle or sentence can be interpreted. By using experts at the first stage to pick out what dimensions of a seed case are important, this process has already narrowed the field of meanings to a small subsection that does not represent groups that are not included among the experts. Pluralism and specification could both be improved by allowing for more complete CBR without the intervention of experts at the first stage and by seeking to use incompletely theorized agreements at the meta-principle level, if such a thing is possible.

*i. Pluralism through diversity of agreements*

The CBR approach has a close relationship with pluralism and its associated concerns because it was at least in part inspired by Sunstein's work on incompletely theorized agreements. Because cases allow for low-level agreement about outcomes while preserving space for disagreements on principle, alignment approaches using cases can theoretically maintain a similar respect for different perspectives on the good while allowing for concrete progress on issues. However, as outlined above, the existing CBR approach does not promote pluralism across a full range of perspectives because it relies on expert determination of the ways in which cases should vary and does not allow for a complete bootstrapping of meaning within the system of cases.

[275] *Id.* The CBR team is inspired by Sunstein's arguments along exactly this axis.
[276] *See* Sunstein, *supra* note 60 at 1735–36.

The CBR approach could be improved by removing the expert determinations piece of the process and replacing it with a method in which principles selected by a process like Collective Constitutional AI are used as the bases for case generation. In this kind of process, users could deliberate about principles at whatever level they deemed best and then those principles would be fed into a LLM that would iterate across them to generate sets of examples of applications of the principles that illustrate edge cases of meaning. So, for example, "fairness" could be chosen as a principle and then the model could produce a bunch of applications of fairness in different contexts. Users would next decide whether they view a given application as actually representing the concept and could write to justify their explanation. This process would allow users to more completely and granularly specify what they mean by the principles that they selected and established "fixed points" from which the aligned model could analogize when applying the principle in future. Cases that implicate multiple principles could be fed to users who had picked one or some of those principles, and situations in which the users who were coming from different principles selected the same answer would effectively demonstrate incompletely theorized agreements that the model could use in future. This variation on the CBR approach would be more completely pluralistic and democratic rather than expert-driven and allow for a more complete and effective specification of the field of meanings that a given principle might have. It would also not be too difficult to apply to the Constitutional AI process.

*ii. Specification by meta-rules and guided examples*

Generally, the existing CBR approach does not allow for the effective specification of meaning across different principles. First, as discussed, it is currently impossible to specify what dimensions of analogy are the important ones. Giving a model enough examples of similar and different cases such that it can figure out those dimensions by seeing where humans change their minds risks lapsing back into the totalizing form of positivism, requiring building case sets that cover almost all of the possibility space for models to learn from, an impracticable task. Second, simple extrapolation is not enough. Whatever dimensions of similarity are chosen should be the subject of democratic deliberation and reasoning if they are to be used to govern people. In some sense, the extrapolations of courts are subject to democratic oversight, at least when they are not in the context of constitutional law. When Congress passes a law to overturn a court decision, it is intervening and telling the court that the extrapolation that it has made is the wrong one to make and specifies how courts should make decisions in that kind of case in the future. Courts can then extrapolate from this kind of intervention when deciding new cases.

CBR alignment needs a kind of meta-level oversight, which should ideally come from a process of deliberation on how models do their analogization into new domains. There is a mathematical literature on features of similarity,[277] formalizing how the presence or absence of features in two cases can provide a measure of the extent to which the cases are similar. Applying that kind of work to law could help provide an overall sense of whether similarity exists across cases and on what that similarity is based with the aim of allowing for better specification of the important dimensions of similarity. The law can also help provide a framework for talking about the relevant features of similarity in extrapolations of new cases. In some sense, this is what the law and legal opinions are doing, explaining why society thinks that certain things are similar and others are different from each other.[278] Indeed, reasoned analogy and distinction is at the core of the legal approach. Sunstein, in his article on AI and analogical reasoning, argued that AI then could not fulfill the necessary functions of legal reasoning because it was unable to identify the correct general principles to guide how extrapolation should be done.[279] Giving models a mix of cases and modes of reasoning about them, in the same way that Sunstein advocates for a mix of precedents and low-level principles as forming the substance of law, would be a useful step forward for the CBR team, and the law should provide a guide for how to build out such a rich set of tools. Finally, as discussed above,[280] increasing the extent to which the public is involved in deliberation about how to think about the similarities and differences between given cases is a crucial step forward for ensuring representative alignment. Public reason-giving and the elaboration of points of consensus and dissensus at both the high and low level would help models learn what kinds of decisions to make in different circumstances and to extrapolate them into the future, moving between these different levels where appropriate in order to provide a flexible and generalizable kind of alignment decisionmaking.

## *IV. Jurisprudence as alignment*

AI models and the tools of alignment will likely be as useful for jurisprudence as jurisprudence is for alignment, but their development and application in the context of law have not been much explored as yet. As such, it is useful to preliminarily investigate some ways that jurisprudence could be improved given AI models of increasing capabilities. Some interesting research is already being done using tools from AI to analyze and evaluate how language works in law, for example by analyzing the semantic similarity of terms used in law as represented in AI models.[281] Judge Newsom of the Eleventh Circuit Court of Appeals recently advocated for using

---

[277] *See, e.g.*, Tversky, *supra* note 61 (providing one influential formalization of the extent to which two things are similar to each other).
[278] If those who believe in analogical reasoning are correct, one implication of their argument is that as much as legal reasoning proceeds by identifying analogies and distinctions between precedents and the case before a judge, legal opinions are, in retrospect, ways of documenting and explaining those distinctions and their justifications.
[279] Sunstein, *supra* note 45 at 33.
[280] *See supra* Part III Section A Subsection ii.
[281] *See* Jonathan H. Choi, *Measuring Clarity in Legal Text*, 91 U. CHI. L. REV. 1, 30–49 (2024) (applying cosine difference measures based on word embeddings to analyze legal problems like the clarity and indeterminacy of text); Yonathan A. Arbel & David A. Hoffman, *Generative Interpretation*, 99 N.Y.U. L. REV. __, 29–43 (forthcoming

AI models to better understand the textual or original meaning of words at issue in cases, training a kind of Founder-LLM that can advise on original constitutional meanings, for example.[282] These improvements in existing modes of analysis within the mainstream of judicial interpretation hold promise for aiding judges in improving their capabilities and making the law more representative of the will of the legislature that has promulgated it.

At a higher level, AI will likely provide two major benefits for the law: first, as just discussed, it will supplement the law, empowering legislators, judges, and lawyers to more effectively describe the world and use law to govern it, and second, it may complement the law by providing a new set of tools and techniques that allow for the accomplishment of the goals and roles of the law even if outside of traditional legal structures. Many of the key objectives of the law revolve around empowering people to more effectively act in the world, regulating other actors and society to open different possibilities for action and coordination. Contract law, for example, allows people to establish mutually-beneficial relationships and interactions without having to trust that the other party has their best interest at heart because the law provides a backstop against misbehavior. The objective of the law is to act as an extender of the capabilities of those who employ and are governed by it, and we can say that the law is "aligned" in a contractual interpretation case to the extent that is does so as those people would have wished at the time that they created the contract.

### A. AI as Supplement to Existing Law

Computers that can reason in natural language and that have a sense of context, while also being increasingly comprehensible via various techniques of interpretability[283] and explainability,[284] raise the possibility of making legal reasoning more transparent and principled.[285] Currently, judgments are susceptible to the critique that they are simply expressions of the policy preferences of the judge taking on the clothes of the rule of law. Such judgments are not aligned to what society would want but instead involve the judge operating beyond her brief to serve her own ends. But society could agree to use a model that analyzed the law based on some set of principles that had been determined by public deliberation to check a judge's reasoning in given cases, constraining this discretion and improving the extent to which the judge is aligned with

2024) (using large language models to analyze the language of contracts and generate predictions about what the parties intended them to mean *ex ante*).

[282] Snell v. United Specialty Ins. Co., 2024 U.S. App. LEXIS 12733 *; _ F.4th _ (11th Cir., 05/28/24) (Newsom, J., concurring).

[283] *See* Linardatos, Papastefanopoulos, & Kotsiantis, *supra* note 30 at 19–20.

[284] *See* Xu et al., *Explainable AI: A Brief Survey on History, Research Areas, Approaches and Challenges*, NLPCC 2019 563, 565-66 (2019).

[285] Nora Belrose & Quentin Pope, *AI is easy to control*, AI OPTIMISM (Nov. 28, 2023), https://optimists.ai/2023/11/28/ai-is-easy-to-control/ (making the argument that AIs are actually relative white boxes compared to the black box of human cognition).

society. An AI trained on the same set of cases that were applied in the judge's decision, and, invoking Dworkin, on related areas of the law,[286] might be used to determine whether the judge's decision and the reasoning behind it are a plausible extrapolation of the precedents being drawn on. Having agreed on a structure of reasoning that could be encoded into the model, using a kind of veil of ignorance[287] about outcomes in order to reason about the principles that guide the model, society could reduce the extent to which judges could simply choose whatever they want and then merely justify those choices after the fact by providing a relatively objective measure of the extent to which the judge is acting out their own preferences. Though Sunstein argued that people are better able to agree on low-level principles and outcomes in particular cases than they are to agree on high-level values,[288] moving debates to the level of values and structure rather than contentious political issues might provide a way for the public to make progress on those hard cases. The obvious final stage of this process is having the evaluator model do the judging of cases itself, but there might be representational or humanistic reasons why we preserve human judges in these roles. In some sense, the AI is acting here as an aligner for the law, applying insights from machine learning and techniques that work on the information processed in that field to provide frameworks for aligning judges more closely with the popular will as expressed through the law.

Another, similar application that aligns the law would be in combining a large language model and a classification algorithm to see whether a given decision is a plausible or correct result. Classification algorithms are a form of supervised learning in AI in which models are given a set of inputs and outputs and learn to map the inputs to the outputs.[289] For example, a model might be given a set of pictures of dogs and cats with labels for whether they are dogs or cats and then learn to classify new pictures into whatever species they are. It seems possible to perform this kind of process in the context of legal reasoning as well. A large language model could be trained on caselaw and would then encode and store those cases in a high dimensional space in its neural network. A classification model could then take in the cases as inputs and their outcomes as outputs and learn to classify when a given new case, encoded in the same way as the precedent cases, would result in a finding of liability or not. When a judge decides a new case, the model could determine whether her decision is consistent with the trends of past cases, providing a kind of objective extrapolation of precedent.

The law already deals with encodings of information in a way that suggests that caselaw is susceptible to this kind of reasoning. When lawyers speak and write to each other, they often speak in a kind of encoding, using the names of cases to stand in for their legal meanings, their holdings.[290] But the encoding is flexible and fuzzy, and, during lawsuits, the meaning of the case will often be disputed through arguments that use the facts of the case to try to give different contents to the holding. This process is at the core of legal reasoning: an earlier decision functions as precedent for the instant case because of some similarities in the facts of each case that form the crux of whether or not to apply the rule from the earlier case. For example, the clear and present danger rule justifying government restrictions of speech[291] is meaningless without a sense of what "clear and present danger" is. That meaning is provided by looking at the facts of the case that supported the holding and at the facts of other cases that were analyzed similarly to *Schenck*, which established that rule. Of course, a model would also need to account for how the law changes, and when it should do so, as the "clear and present danger" test was modified to the now-governing "imminent lawless action" test in *Brandenburg v. Ohio*.[292] The new test supplants the old but still bears some relation to it—the fact that it was deemed necessary to replace the old test in fact gives us information about certain relevant features of free speech law in changing times. On each issue, cases form a web of decisions in relation to each other, some in which the court held one way and some in which the court held the other, and as Sunstein argued, the core problem is figuring out which similarities and which differences are dispositive.[293] By understanding how models reason about particular cases and make analogies and distinctions, it seems plausible that we could begin to better see how humans do so too. AI could provide a more objective way of answering that question because models would not be motivated by getting to particular political outcomes when analyzing what connects and explains past cases.[294]

[286] Dworkin, *supra* note 3 at 245.
[287] *See generally*, JOHN RAWLS, A THEORY OF JUSTICE (1971).
[288] Sunstein, *supra* note 59 at 1735–36.
[289] *See* Pratap Chandra Sen, Mahimarnab Hajra, & Mitadru Ghosh, *Supervised Classification Algorithms in Machine Learning: A Survey and Review*, *in* EMERGING TECHNOLOGY IN MODELLING AND GRAPHICS 99, 100 (Jyotsna Kumar Mandal & Debika Bhattacharya eds., 2020)

Finally, AI might provide a way out of the trap of the meaning of language, or at least help us make progress in it. Legal theory confronts the problem of the limits of language; it is fine for Dworkin to say that judges should decide cases at least in part according to a theory of justice, but unless we can come to some agreement about what justice means, something that has long escaped philosophy, then ultimately his command is empty. That realization is at the core of the positivist critique of interpretivism: at least statements of positive law provide some ground for interpreting what the political community wants to do, while abstract values are difficult or impossible to consistently apply.[295] But positivists are not immune to this critique, and similarly, Sunstein does not ever fully answer the problem of how to determine which similarities and differences among cases are the important ones, something that he acknowledges is an essential part of a theory of analogical reasoning.[296] Instead, in his paper on analogical reasoning and AI, Sunstein sounds downright Dworkinian, writing that analogical reasoners must decide on what principle of analogy "is actually *better*,"[297] that is to say makes "best constructive sense out of a past decision."[298] Sunstein suggests that this quality of "betterness" relates to what is best for the overall welfare,[299] but in doing so seems to betray his arguments for incompletely theorized agreements, which surely cannot include an agreement in favor of utilitarianism.[300] Hart acknowledged the same problem, and explicitly argued that it was a feature of language that could likely not be overcome.[301] As noted above, alignment currently faces a similar problem.[302] Principles like "good for humanity" are so vague as to be effectively meaningless, except that they seem to have meaning for large language models, which can use them to make more helpful and safer decisions.[303] Large language models understand the meanings of words through the public use of them, apparently able to participate in "what Wittgenstein called a form of life sufficiently concrete so that the one can recognize sense and purpose in what the other says and does."[304] Wittgenstein wrote, "to imagine a language is to imagine a form of life,"[305] and it seems clear that these models are engaging in the form of life that is the law. The budding sciences of interpretability and explainability might help open the black box of AI and, in doing so, crack open the black box of the law, allowing us to understand better what the law is and how lawyers and judges reason, and should reason, within it in order to make them more aligned with what society would prefer.

---

[290] This practice is a commonplace, and cases occasionally take on meanings that go beyond their actual facts or holdings, taking on symbolic meanings.
[291] *See* Schenck v. United States, 249 U.S. 47, 52 (1919).
[292] Brandenburg v. Ohio, 395 US 444, 447 (1969).
[293] Sunstein, *supra* note 58 at 774.
[294] *Cf.* Snell, 2024 U.S. App. LEXIS 12733 *37 (Newsom, J., concurring) (arguing that LLMs might be more reliable sources of meaning than humans because they do not rely on manipulated inputs or have political objectives).
[295] *See* Leiter, *supra* note 14 at 16.
[296] Sunstein, *supra* note 58 at 774.

### B. AI as Complement Beyond Existing Law

Beyond providing a framework for the alignment of existing legal systems to the preferences of society, AI systems will likely also provide new quasi-legal tools for people to use to accomplish many of the goals that the legal system currently enables. The recent rise of agentic AI systems[306] and the legal[307] and multi-agent alignment[308] problems that they present provides a useful demonstration of how this kind of complementation process might play out. Agentic AIs act on behalf of the user, performing various tasks according to the user's statements of what they want. Leading current versions of these systems include OpenAI's Operator[309] and Deep Research[310] systems, as well as the Manus system that went viral in the early spring of 2025.[311] In each of these systems, the user provides a command phrased in natural language, and then the system goes onto the internet and takes actions in accordance with the command of the user. The set of tasks that these systems can complete is relatively limited at present, and they are unable to accomplish tasks that require working over a long time horizon, but both of these restrictions on utility are being solved.[312] Furthermore, some of these systems are able to check in with the user when they face a situation that they are unable to resolve, like having to enter credit card information or make a choice between different ways to proceed. This checking-in function demonstrates the beginnings of a kind of contextual understanding of situations of ambiguity that would be very useful for AI systems across a variety of domains, including legal ones as discussed above.

---

297 Sunstein, *supra* note 60 at 33 (emphasis original).
298 *Id.* (citing Dworkin, *supra* note 3 at 67–68).
299 *Id.*
300 Sunstein, *supra* note 59 at 1738 (citing utilitarianism as an example of a "general theory," contrasted with incompletely theorized agreements).
301 Hart, *supra* note 1 at 127-28.
302 *See supra* Part III Section A.
303 Kundu et al., *supra* note 112 at 24.
304 Dworkin, *supra* note 3 at 63 (citing Wittgenstein, though it is unclear from the text what the exact source of the paraphrase is).
305 LUDWIG WITTGENSTEIN, PHILOSOPHICAL INVESTIGATIONS 8e (1953).
306 *See* Kolt, *supra* note 15.
307 *Id.*
308 *See* Maha Riad, Vinicius Renan de Carvalho, & Fatemeh Golpayegani, *Multi-Value Alignment in Normative Multi-Agent System: Evolutionary Optimisation Approach*, ARXIV (May 12, 2023), https://arxiv.org/abs/2305.07366; Edmund Dable-Heath, Boyko Vodenicharski, James Bishop, *On Corrigibility and Alignment in Multi Agent Games*, ARXIV (Jan. 9, 2025), https://arxiv.org/abs/2501.05360v1.

More significantly, these AI systems may at some point be able to act in quasi-legal ways, operating like legal agents on behalf of their users and empowering them to accomplish their goals in ways mixed with and similar to the ways that the law does so. For example, the capacity to contract on behalf of the user, entering into legally-enforceable relationships that accomplish goals that the user might have, would shift many burdens from people onto AI systems that are aiding them, especially if they become better at contracting than the average person. An AI could be aligned to the preferences of a user, having both a rich sense of those preferences and the ability and knowledge to check in with the user in situations in which the original commands of the user were not sufficient to unambiguously guide the agent in this novel situation. Then, the agent could act on the user's behalf to accomplish a wide variety of tasks, much broader than the normal set that people contract over because of the high burdens of contracting. Imagine that a user is surfing the internet and their attention is being harvested through advertising cookies. The value of visiting a given website to the advertiser is probably currently measured in some tiny amount based on the likelihood that the visit converts into a sale, but companies like Google and Meta have amassed great fortunes from the margins that they get from directing the attention of users to advertisers.[313] Users cannot contract over such tiny amounts per website, leading to relatively ineffective broad-brush solutions like the cookie consent requirements of the GDPR.[314] If, instead of legislatures having to take that kind of measure to respond to the diffuse but substantial economic effects of advertising, AI agents could simply contract on behalf of users for small payments for visiting given websites, users might be able to capture some of the economic benefits of the direction of attention. The economics of this kind of solution would have to be worked out and might not work depending on the inference cost of automated systems contracting with each other, but it points to ways in which AI systems could empower people to use the tools of the law by decreasing the cost of them accessing those tools. AI agents that helped people buy houses could work on a similar model and would in many cases likely improve the extent to which people were able to accomplish their goals providing cheap expertise in negotiations.

---

[309] *Introducing Operator*, OPENAI (Jan 25, 2025), https://openai.com/index/introducing-operator/.
[310] *Introducing deep research*, OPENAI (Feb. 2, 2025), https://openai.com/index/introducing-deep-research/.
[311] Caiwei Chen, *Everyone in AI is talking about Manus. We put it to the test.*, MIT TECH. REV. (Mar. 11, 2025), https://www.technologyreview.com/2025/03/11/1113133/manus-ai-review/.
[312] *See*, *Measuring AI Ability to Complete Long Tasks*, METR (Mar. 19, 2025), https://metr.org/blog/2025-03-19-measuring-ai-ability-to-complete-long-tasks/.
[313] *See How our business works*, GOOGLE (n.d.), https://about.google/intl/ALL_uk/how-our-business-works/; Matthew Johnston, *How Does Facebook (Meta) Make Money?*, INVESTOPEDIA (Jun. 29, 2024), https://www.investopedia.com/ask/answers/120114/how-does-facebook-fb-make-money.
[314] *See* Christine Utz et al., *(Un)informed Consent: Studying GDPR Consent Notices in the Field*, ACM CCS '19, https://dl.acm.org/doi/pdf/10.1145/3319535.3354212.

In public life, aligned AI systems could represent people in public deliberations and lawmaking processes that affect their lives but that they lack the capacity to deal with themselves. For example, many local government proceedings have significant implications for the people who live in cities around the country, but those people lack the time, attention, or resources to engage regularly in such events. Having AI agents who know the preferences of their users who can attend and participate in such processes on their behalf could significantly increase the extent to which people are able to have their views and preferences represented through public deliberative processes. This kind of representative AI could be scaled up to state and even federal deliberations, reducing the extent to which people have to rely on another kind of agent, representative lawmakers, and narrowing the gap between their desires and what is made into law. Reducing the frictions of the lawmaking process and decreasing the costs of participation by enabling people to offload certain parts of the work that goes into such participation could make society and government more broadly democratic. Even if such offloading were only partial, such that AIs notified people of particularly relevant issues for them that were the subject of lawmaking, rather than complete representative replacement of human participation, there would likely be substantial benefits to democratic engagement in lawmaking. Such a process would make the law as a whole more aligned with the actual interests of people in society rather than having that alignment be only to what the representatives believed those interests to be, as mediated by occasional elections. While we might not want to accept full replacement of human participation in decisionmaking with AI participation, even if those AIs are robustly aligned to the interests of their human users, it is worth considering how AI could be more effectively incorporated into government in this kind of way.

Finally, AI systems may soon exceed human abilities across a wide variety of cognitive tasks.[315] This kind of artificial general intelligence (AGI) may be able to do better for their users than the users themselves would know how to do, both in using the law and in using other social technologies. If so, there will be significant pressures to replace human decisionmaking with AI decisionmaking in different parts of the economy and society because the outcomes of allowing the AI to make decisions will be better from many perspectives than retaining human participation.[316] Replacing human workers with higher-performing and cheaper AIs is one clear example of this kind of process, illustrating how even companies that might prefer to retain human employees could be forced by competitive pressures to replace them in order to stay competitive with other firms that have become more automated. Military competition and the pressures of security dilemma situations presents another useful example.[317] If AIs become better than humans at different parts of governing, for example in running regulatory programs or the Federal Reserve, or in distributing social welfare benefits, then the question of the extent to which we prioritize preserving human participation in these activities at the price of them being carried out less effectively, will become a real one. The law will become a key mechanism for ensuring that any commitments to preserving human participation in key decisionmaking processes can survive competitive pressures to replace people, and ensuring that where AI systems do replace humans, they are aligned to those on whose behalf they are acting and to humanity more generally will become essential.

---

[315] Leading AI developers and researchers tend to think that AGI will arrive within the next five or so years. Lakshmi Varanasi, *Here's how far we are from AGI, according to the people developing it*, BUSINESS INSIDER (Nov. 9, 2024), https://www.businessinsider.com/agi-predictions-sam-altman-dario-amodei-geoffrey-hinton-demis-hassabis-2024-11.

## *Conclusion*

Law and alignment are isomorphic—they have the same shape, and operations performed on one may well be applicable to the other. These two bodies of research, the one in legal theory and the other in AI, can usefully enter into conversation together. This paper has sought to begin that conversation, identifying similarities between significant legal theories and alignment approaches and showing how each field can help solve the problems of the other. In particular, both fields must ensure democratic representation and create predictable and functional constraints on the actions of powerful decisionmakers that apply in novel situations. The two sets of approaches, one oriented at inculcating values and general principles and the other in defining through example, each represent a promising path forward, but one that is also marked with difficulties. Combining them, by using meta-principles of extrapolation given content by particular case-based examples of those kinds of extrapolation, is likely the best path forward. This mixed approach best allows for the specification of how models should decide in the future by using the strengths and limits of language to create a precise but flexible mode of alignment. The law may help solve the alignment problem. On the other side, AI may help advance jurisprudence by providing a set of tools for better analyzing what the law is, and perhaps what it should be. The introduction of general foundation models, machines that can reason in language and in context, marks a shift in how reasoning is done, in the law and outside of it.

In a prescient article from 2001, Sunstein argued that AI was then unable to perform legal reasoning because, even if it could assemble sets of cases and suggest similarities and differences among them that could be applied to resolve new situations, it was unable to engage with legal reasoning's "inevitably evaluative, value-driven character," unable to reason about the "principles" that make one case more or less like another.[318] Quoting,[319] and sounding like, Dworkin, Sunstein argued that the closeness of a precedent to the instant case depends "on identification of a (normative) principle by which 'closeness' can be established."[320] Yet Sunstein concluded that this failure of AI might one day be remedied, and "computer programs . . . . engage with [legal reasoning] on their own."[321] That day may have come. New models can reason effectively in law, and alignment seeks to provide exactly the kind of normative evaluative principles that Sunstein argued were lacking in 2001. If that day has arrived, we will see great changes as models take on roles of power and decision heretofore restricted to humans. It is necessary to begin thinking more deeply about the ways that the law and AI relate.

---

[316] *See* Jan Kulveit et al., *Gradual Disempowerment: Systemic Existential Risks from Incremental AI Development*, ARXIV (Jan. 28, 2025), https://arxiv.org/abs/2501.16946.
[317] *Id.*

[318] Sunstein, *supra* note 60 at 31.
[319] *Id.* at 32 (quoting Ronald Dworkin, *In Praise of Theory*, 29 ARIZ. ST. L.J. 353, 371 (1997)).
[320] *Id.* at 33.
[321] *Id.* at 35.